\documentclass[10pt,twocolumn,letterpaper]{article}

\usepackage{cvpr}              

\usepackage[dvipsnames]{xcolor}

\usepackage[utf8]{inputenc} 
\usepackage[T1]{fontenc}    
\usepackage{url}            
\usepackage{booktabs}       
\usepackage{amsfonts}       
\usepackage{nicefrac}       
\usepackage{microtype}      
\usepackage{xcolor}         
\usepackage{scalerel}

\usepackage{times}
\usepackage{epsfig}
\usepackage{graphicx}
\usepackage{amsmath}
\usepackage{amssymb}
\usepackage{array,multirow}

\usepackage{graphicx}
\usepackage{amsmath}
\usepackage{amssymb}
\usepackage{booktabs}

\usepackage{rotating}

\usepackage{tabularx}
\usepackage{colortbl}
\usepackage{bm}
\usepackage{xspace}

\usepackage{enumitem}
\usepackage[olditem,oldenum]{paralist}

\usepackage[belowskip=1pt,aboveskip=5pt,font=small]{caption}
\usepackage[belowskip=0pt,aboveskip=3pt,font=small]{subcaption}
\newcommand{\xhdr}[1]{\vspace{3pt}\noindent\textbf{#1}\xspace}

\newcolumntype{s}{>{\columncolor[gray]{.85}[.5\tabcolsep]}c}

\makeatletter
\DeclareRobustCommand\onedot{\futurelet\@let@token\@onedot}
\def\@onedot{\ifx\@let@token.\else.\null\fi\xspace}

\def\eg{\emph{e.g}\onedot}
\def\vs{\emph{vs}\onedot}

\usepackage[belowskip=1pt,aboveskip=5pt,font=small]{caption}
\usepackage[belowskip=0pt,aboveskip=3pt,font=small]{subcaption}
\newcommand{\vince}[1]{\textcolor{blue}{[VC:#1]}}

\makeatletter
\DeclareRobustCommand\onedot{\futurelet\@let@token\@onedot}
\def\@onedot{\ifx\@let@token.\else.\null\fi\xspace}

\def\eg{\emph{e.g}\onedot}
\def\vs{\emph{vs}\onedot}

\definecolor{cvprblue}{rgb}{0.21,0.49,0.74}
\usepackage[pagebackref,breaklinks,colorlinks,citecolor=cvprblue]{hyperref}

\usepackage[capitalize]{cleveref}
\crefname{section}{Sec.}{Secs.}
\Crefname{section}{Section}{Sections}
\Crefname{table}{Table}{Tables}
\crefname{table}{Tab.}{Tabs.}

\def\paperID{13137000000} 
\def\confName{CVPR}
\def\confYear{2024}

\title{3D Semantic MapNet: Building Maps for Multi-Object Re-Identification in 3D}

\author{
    Vincent Cartillier$^1$, Neha Jain$^{2}$, Irfan Essa$^{1,3}$ \\
    $^1$GeorgiaTech \quad $^2$ Meta \quad$^3$ Google \\
    {\tt\small vcartillier3@gatech.edu} \\
    \tt\normalsize
    \href{https://vincentcartillier.github.io/3d_smnet.html}{vincentcartillier.github.io/3d\_smnet.html}
}

\begin{document}
\maketitle

\begin{abstract}
We study the task of 3D multi-object re-identification from embodied tours. 
Specifically, an agent is given two tours of an environment (\eg an apartment) under two different layouts (\eg arrangements of furniture). 
Its task is to detect and re-identify objects in 3D  -- \eg a `sofa' moved from location A to B,
a new `chair' in the second layout at location C, or a `lamp' from location D in the first layout missing in the second. 
To support this task, we create an automated infrastructure to generate paired egocentric tours of initial/modified layouts 
in the Habitat simulator \cite{habitat19iccv,szot2021habitat} using Matterport3D scenes \cite{chang2017matterport3d}, 
YCB \cite{calli2015ycb} and Google-scanned objects \cite{google_scans}. 
We present 3D Semantic MapNet (3D-SMNet) -- a two-stage re-identification model consisting of 
(1) a 3D object detector that operates on RGB-D videos with known pose, and 
(2) a differentiable object matching module that solves correspondence estimation between two sets of 3D bounding boxes. 
Overall, 3D-SMNet builds object-based maps of each layout and then uses a differentiable matcher to re-identify objects across the tours. 
After training 3D-SMNet on our generated episodes, we demonstrate zero-shot transfer to real-world rearrangement scenarios by instantiating our task in Replica \cite{replica19arxiv}, Active Vision \cite{ammirato2017dataset}, and RIO \cite{wald2019rio} environments depicting rearrangements. On all datasets, we find 3D-SMNet outperforms competitive baselines. Further, we show jointly training on real and generated episodes can lead to significant improvements over training on real data alone.

\end{abstract}    
\section{Introduction}

\begin{figure}[t]
    \centering
    \includegraphics[width=1\linewidth]{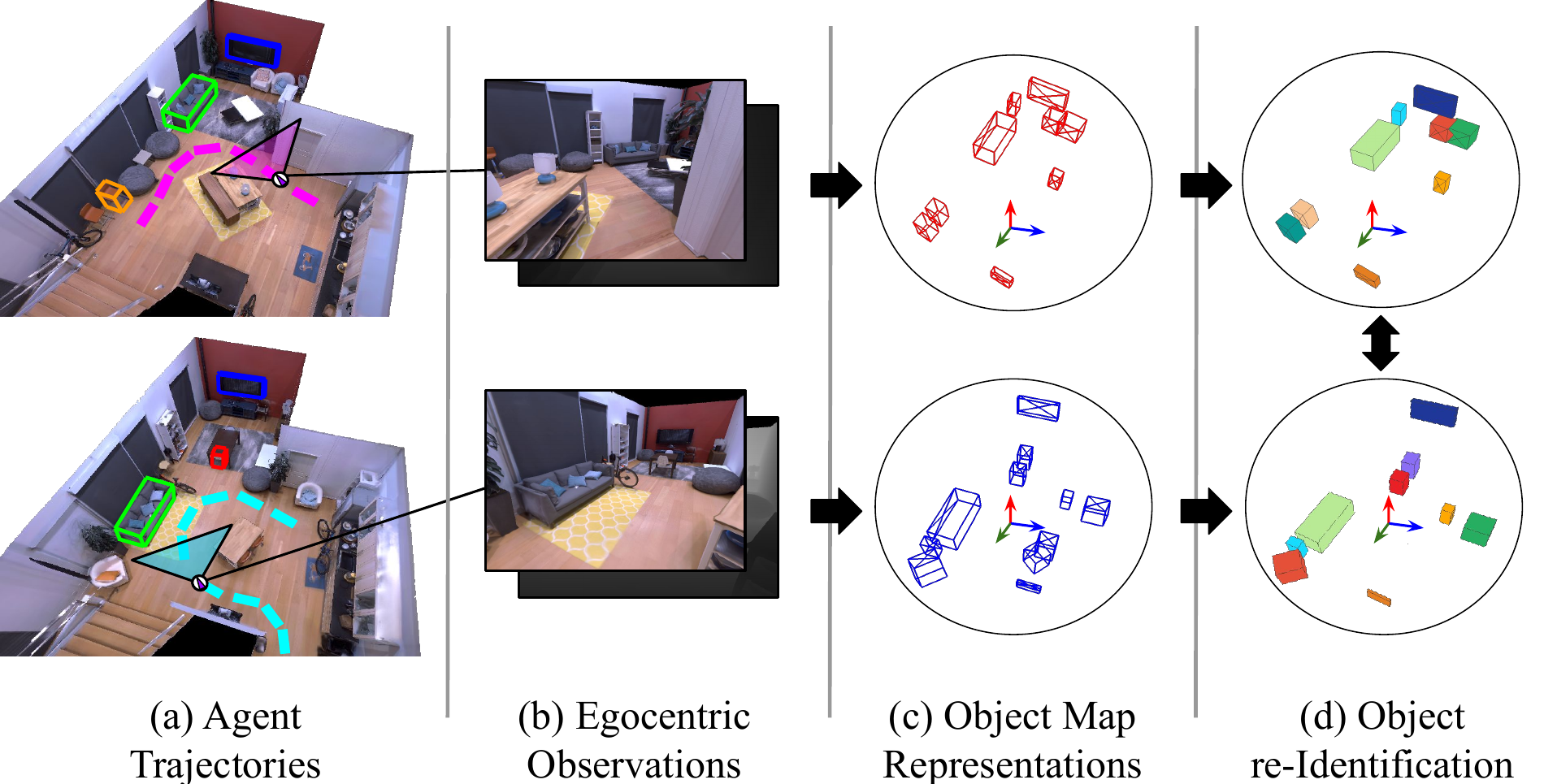}\\[5pt]
    \caption{3D Multi-Object Re-Identification: an agent is provided two tours of an environment (egocentric RGD-D videos with known pose). 
    The two layouts may differ with objects added (red), removed (orange), moved (green) or unchanged (blue). 
    The goal for the agent is to detect and re-identify objects in 3D. 
   } 
\label{fig:teaser}
\vspace{-15pt}
\end{figure}
Imagine a home assistant robot asked to tidy up a house after an event or a party. Such an agent needs to have complete context of what objects belong where, what was added, and is missing. It also needs to know where all the objects are now, were before, and create a map of how the scene has changed in order to restore it back to its normal state.  This example illustrates a broadly useful skill for embodied agents interacting with human environments -- the ability to represent the world as a set of dynamic objects that persist over time. This is foundational to linking user commands to entities in the world that are not currently in the field of view, answering questions about the location of objects (\eg where did I leave my phone?), and reasoning about likely events that have changed the state of the world (\eg someone must be home because a new coat is on the rack).

Within the context of this longer-term goal, we study the task of 3D multi-object detection and re-identification in an indoor space. An embodied agent (a virtual robot or an egocentric AI assistant) is equipped with an RGB-D camera with known pose (extracted via a localization system). The agent is provided two tours (represented as camera trajectories) of an environment under different layouts. Between the two layouts, objects may have been added, removed,  moved, or be unchanged. The agent needs to detect 3D bounding boxes and match them between the two layouts (as illustrated in \cref{fig:teaser}).
This is a challenging and realistic problem that goes beyond the classical 2D or 3D object detection formulation.  
First, the agent must localize objects in 3D in varied poses and contexts from 
passive egocentric views (it does not control how an object is framed). 
Second, it must match instances across layouts 
despite large changes in pose or relationships to surrounding objects 
(\eg a lamp may be moved across the room, breaking inter-object contextual relationships). 
Third, object instances across tours may have differing degree of observation (\ie an object seen from multiple sides in the first tour might only be seen from one in the second).

Our approach, called 3D Semantic MapNet (3D-SMNet), consists of two natural components following the structure of the task --
\begin{compactenum}[(1)]
\item \textbf{a 3D object detector} that takes as input an RGB-D video with known poses and outputs an object-based map. The map (\cref{fig:teaser}(c)) stores a set of object instances, each defined by its 3D bounding-box centroid, orientation and dimensions, class label, and a feature descriptor used for the re-identification phase. The object-map for each layout are then passed to 

\item \textbf{a matching module} that estimates correspondences between two sets of 3D bounding boxes, based on appearance, geometry, 
and context (\cref{fig:teaser}(d)). 
\end{compactenum}
To benchmark the current state of the field on this task, we take a straight-forward approach of instantiating these components with recent prior works as (1) a VoteNet\cite{qi2019deep} detector and (2) a modified SuperGlue\cite{sarlin2020superglue} optimal matching layer. To improve performance on added and removed objects, we introduce a novel scene-adaptive mechanism for identifying unmatchable objects.


We conduct experiments using photo-realistic scans of building-scale environments (homes, offices, churches) in the Matterport3D (MP3D) dataset~\cite{chang2017matterport3d} using the Habitat simulation platform~\cite{habitat19iccv,szot2021habitat}, which allows access to agent state, navigation trajectories, RGB-D renderings, \etc. The scans in MP3D are `monolithic', \ie the entire environment is stored as a single textured mesh and individual objects can not be moved or removed. To overcome this problem, we use the Google-Scanned-Objects (GSO) \cite{google_scans} and YCB objects \cite{calli2015ycb} datasets, which cumulatively consist of over 1,000 3D scans of objects from different semantic categories that can be inserted in MP3D environments. 
We create an automated infrastructure to generate paired egocentric tours of initial/modified layouts of MP3D scenes populated with different objects from GSO and YCB; we expect this infrastructure to be broadly useful beyond this task. We conduct extensive experiments in this generated dataset to quantify the impact of model choices in 3DSMNet against competitive baselines.

To further demonstrate the usefulness of our approach and data generation infrastructure, we show zero-shot transfer results to scans of real environment rearrangements from the Replica \cite{replica19arxiv}, Active Vision \cite{ammirato2017dataset}, and RIO \cite{wald2019rio} datasets. We adapt our task to these environments and show 3D-SMNet outperforms competitive baselines. Finally, we show that jointly training on real and generated data results in significant improvements for re-identification in the RIO environments -- improving matching accuracy by $+10\%$ on real environments compared to when the system is trained on real data alone.

\section{Related work}

\xhdr{Re-identification} has been extensively studied for class-specific problems such as person re-ID \cite{ye2021deep,wieczorek2021unreasonable,ni2021flipreid,sharma2021person} or vehicle re-ID \cite{wang2019survey,huynh2021strong,zheng2020vehiclenet,he2021transreid}. 
Typically, a system is provided a query image and asked to rank a list of gallery images by similarity. 
In contrast, we solve a matching 
problem with 3D detections (not a ranking problem).
In addition, most existing works use 2D CNN \cite{wieczorek2021unreasonable,ni2021flipreid,huynh2021strong,zheng2020vehiclenet} or 
2D transformer features \cite{sharma2021person,he2021transreid}. We describe objects with 3D features.
In person re-ID, some works \cite{baltieri2011sarc3d,zheng2020person,chen2021learning} do perform re-identification in 3D, however these approaches 
leverage a body model; we assume no access to 3D models.

\xhdr{3D Multi-Object Tracking (MOT).} 
Most multi-object trackers have a data association module used to match tracklets with detections \cite{weng20203d,yin2021center,chiu2020probabilistic}. The score matrices associated with these matchers are computed using proximity metrics such as 3D-IoU \cite{weng20203d}, Euclidean distance \cite{yin2021center} or Mahalanobis distance \cite{chiu2020probabilistic}.
This differs from our problem as we are trying to match objects based on their appearance without any notion of physical distance.

In 2D-MOT, many methods include a `re-identification module' to handle long-term occlusions \cite{al2018multi,bergmann2019tracking,tang2017multiple}. 
Unlike our work, these methods use 2D detections and features. 
Closest to our work is ODAM~\cite{li2021odam}, where objects are detected and associated using a learned matcher based on 2D detections. 
A key difference is that in contrast to ODAM, our approach aims to re-identify objects in 3D that have been moved. 
To the best of our knowledge, we are not aware of of any prior work on multi-class object re-identification of 3D detections, which is our focus.

\xhdr{Indoor multi-class 3D object re-identification.}
Replica \cite{replica19arxiv}, ReplicaCAD \cite{szot2021habitat}, RIO \cite{wald2019rio} and Active Vision \cite{ammirato2018active} are interesting datasets to support indoor 3D object re-ID. 
They all contain house meshes under different layouts. 
However, Replica and Active Vision are too small to support training and the object models in ReplicaCAD are the same across all scenes which 
does not allow for testing generalization to unseen instances. The RIO dataset consists of several RGB-D scans semantically annotated of houses. We use the RIO dataset for training and testing. We also report results on the Replica and Active Vision datasets.

Multiple datasets from the robotic object rearrangement domain exist~\cite{danielczuk2021object,huang2019large}. However, these datasets mostly focus on small scenes (\ie tabletop scenarios). Recently, object re-arrangement tasks and datasets for Embodied AI domains have appeared~\cite{szot2021habitat,batra2020rearrangement}, including efforts~\cite{szot2021habitat} that provide pairs of large scenes with moved objects. However, the differences between scenes is usually limited to a few objects from the YCB dataset~\cite{calli2015ycb}. In our task we seek to scale the number of scene differences by a factor of $10$. 

The Robotic Vision challenge \cite{hall2020robotic} presents two tasks, Semantic SLAM and Scene Change Detection (SCD) which are closely related to our problem. In Semantic SLAM the goal is to explore an environment and build an object-based map of the scene.
Three baselines have been submitted to this challenge: two of them build object-based maps of a point-cloud based 3D object detector (VoteNet) \cite{qi2019deep} and the third one gets 3D detections from aggregated 2D depth segmentations (via RGB-segmentations using MaskRCNN \cite{he2017mask}). Our approach to construct an object-based map from one exploration is similar to the first set of approaches and uses VoteNet \cite{qi2019deep} as well.
In the SCD Challenge, an agent explores an environment at two different times, its goal is to output an object-based map of the scene with objects being added or removed from the environment -- note it does not consider movement of objects.
At the moment, no baseline or submission has been reported for this task. 



\section{Multi-Object Re-identification from  Tours}
To recap, we consider a multi-object re-identification problem in 3D environments observed through egocentric tours. 
Each problem instance (or episode) is defined by a pair of egocentric tours through different layouts of the same environment 
-- an initial layout and a modified layout. 
Objects in the initial layout may be moved, removed, or unchanged in the modified layout. 
Further, new objects may be added to the modified layout. 
The task is to re-identify objects present in both layouts and identify objects which have been removed or added. 
Notably, we do not constrict the two tours to follow the same path. 

To support this problem definition, we develop a procedure to generate paired egocentric tours of initial/modified layouts 
in the Habitat simulator \cite{habitat19iccv,szot2021habitat}. 
At a high level, we use Matterport3D (MP3D) \cite{chang2017matterport3d}  environments and insert YCB \cite{calli2015ycb} and Google-scanned-objects \cite{google_scans} to create initial and modified layouts. 

\subsection{Environments and Objects}
The MP3D dataset contains 3D meshes of (mostly) home environments that on average have 517.34$m^2$ of floorspace. 
These meshes also have semantic labels and we retain five of the MP3D classes as part of our task: \emph{chair, bed, toilet, couch, and potted plant}. Similar to \cite{chaplot2020semantic}, we chose these as they overlap with COCO classes \cite{lin2014microsoft} and correspond to common objects in indoor scenes. However, objects in MP3D are part of monolithic scene mesh and cannot be moved or removed. 

Thus, we insert, move, and remove 3D object models from the YCB \cite{calli2015ycb} and Google-Scanned-Objects (GSO) \cite{google_scans} datasets. We manually group GSO objects into semantic categories and then select 10 categories from GSO and YCB: \emph{bag, figurine, plant, puzzle-toy, vehicle-toy, lotion, puzzle-toy, doll-toy, cartridge, dietary-supplement, shoe}. Each object category consists of multiple 3D models. In total, we use 632 models split into train/val/test such that no 3D model repeats across splits. 
We scale the inserted 3D models up to unrealistically sizes to improve their visibility in the egocentric views and to align with what 
modern 3D object detectors are capable of detecting. 
This assumption is a reflection of the state of art and should be relaxed in the future as progress is made.

\subsection{Episode Creation}
Episode creation is a three-step process: creating the initial layout, modifying that layout, and generating egocentric tours in both to serve as input.

\xhdr{Initial Layouts.} To generate an initial layout, we sample a floor in a MP3D environment and then insert 30 objects. Objects models are sampled by class and then instance. The sampled model is inserted in the environment at a random location on either the floor or the top of a `receptacle' support surface.  
Receptacles include tables, sofas, counters, and beds and are manually annotated with a support surface. 
Objects are inserted `upright', \ie with their bounding boxes gravity-aligned, and with random rotation about the vertical axis. And then a collision detector is used to check if the insertion was successful. If collisions exist, the sample is rejected and we repeat this sampling procedure.

\begin{figure*}[t]
    \centering
    \includegraphics[width=0.8\linewidth]{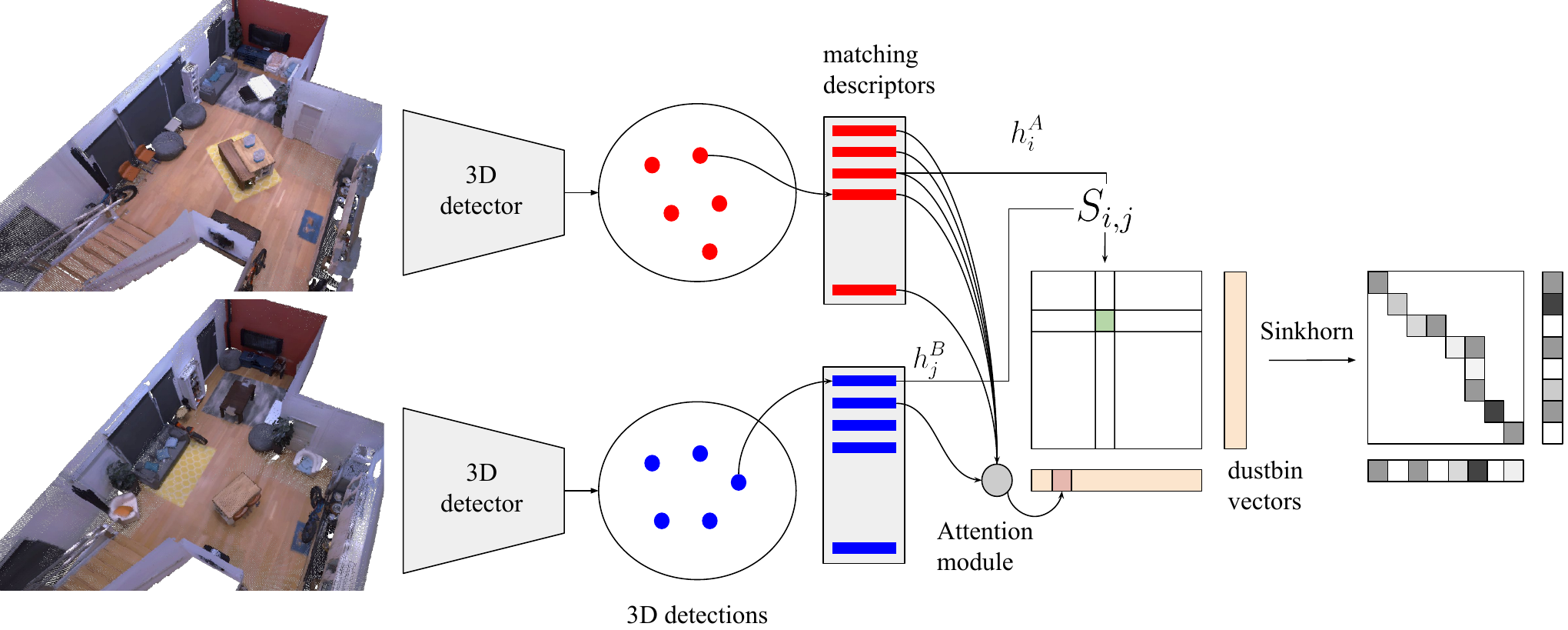}
    \caption{3D-SMNet consists of a 3D object detector and a matching module. The 3D object detector~\cite{qi2019deep} takes as input a textured point-cloud representation of the scene and outputs a set of 3D detections along with feature descriptors. The matching module computes similarity scores from the pairwise descriptors and then extends the score matrix with dustbin vectors estimated from an attention mechanism over the two sets of features to capture added/removed objects. The Sinkhorn algorithm \cite{sinkhorn1967concerning} is then applied to solve the partial assignment problem.}
    \label{fig:3d_smnet_model}
\vspace{-10pt}
\end{figure*}

\xhdr{Modified Layouts.}
To create modified layouts, we move and remove inserted objects from the initial layout and add new ones. Specifically, we leave 1/3rd in the same position, move 1/3rd to new locations, and remove 1/3rd then add an equal number of new objects. New locations for moved objects and inserted objects are sampled identically to the procedure described above. This corresponds to modified scenes with 30 objects -- 10 moved, 10 unmoved, and 10 new objects. 
This translates to 40 objects observed across the tours with 20 being re-identifiable and 20 being removed or added. 
We randomly rotate and translate the 3D environment so models have to reason about relative arrangement and visual appearance rather than absolute position to recognize unmoved objects.

\xhdr{Egocentric Tours.}
We develop an iterative sampling procedure to build trajectories and then run a simulated agent through the trajectory to collect an RGB-D tour. 
A set of $\approx$ 1000 locations are densely sampled in the environment such that each is at least a meter from any obstacle. 
One of these locations is sampled as the start location and then the trajectory extends by sampling the next closest location at least 2 meters away of the current location. A shortest-path planner is used connect the locations. Once a location has been selected, it is removed from the candidate set along with any location within 2 meters. The tour is ended when no next location is available. While this does not ensure full coverage of all objects, we find the paths cover nearly the entire environment.

\xhdr{Dataset Statistics.} 
Following the strategy described, we create 625 episodes split 461/65/126 between train/val/test. This corresponds to a total of 24k, 3k, 7k unique object pairs. Tours consist of 800 steps on average. It provides good space coverage with $78\%$ of objects being actually observed during the tours.

\section{3D Semantic MapNet (3D-SMNet)}


Our approach, named 3D Semantic MapNet (3D-SMNet) and illustrated in \cref{fig:3d_smnet_model}, consists of two broad components: 
(1) a 3D object detector that operates on RGB-D videos with known pose, and 
(2) an object matching module that solves correspondence estimation between two sets of 3D bounding boxes. 

\subsection{3D object detector} \label{3d_object_detector}
We experiment with the VoteNet~\cite{qi2019deep} object detector. VoteNet is a point-cloud based 3D object detector. 
We create the point-cloud inputs for each layout by un-projecting the depth pixels at each position along the camera trajectory using a pinhole camera model. 
We accumulate all unprojected pixels of all frames accross the trajectory to generate a point-cloud $P$ with $N'\times 3$ points. To avoid ceiling detections we tilt the camera by a $22.5$ deg rotation looking downwards.  
Additionally, during the unprojection step, we ignore 3D points un-projected $50$ cm above the camera height.
We further preprocess the input by subsampling the accumulated point-cloud $P$ using a voxel resolution of $3$ cm to only keep $N$ points in the point-cloud. Following this process results in an average input dimension of 2M points.


\xhdr{VoteNet training.} We train VoteNet on the 3D-SMNet episodes (Matterport scenes \cite{chang2017matterport3d} with YCB \cite{calli2015ycb} and Google-scanned objects \cite{google_scans}). Considering the very large point-cloud, we train VoteNet on crops of the full point cloud. We create a training set by selecting $4\times3\times4$m crop regions of the accumulated point cloud. We experimentally found that selecting training crops centered on objects of interest leads to better generalisation results. Please refer to the supplement for training details and additional results.

\xhdr{VoteNet inference.} At inference time, we use a sliding window approach and apply the trained VoteNet module on small point-cloud crops of dimensions $4\times3\times4$m with a stride of $2m$, $1.5m$, and $2m$ in the $x, y,$ and $z$ directions (recall that the $y$-axis is the gravity-aligned axis). We aggregate all the generated object proposals and then apply non-maximum suppression (NMS) to remove duplicates. 
We found that setting a confidence threshold of $0.99$ with a NMS-3D-IoU threshold of $0.05$ led to the best object-detection 
performance with a 
mean average precision (at 0.25 IoU threshold) $mAP@25$ of $0.555$, $mAP@50$ of $0.369$, and an 
Object Map Quality ($OMQ$) \cite{hall2020robotic} of $0.14$.

\xhdr{Object-based maps.}
 We define an object-based map $(M)$ as a set of $m$ objects such that $M = \{O_1,...,O_m\}$. 
 Each object $(O)$ is defined by its bounding-box centroid $c=[c_x,c_y,c_z]$ and dimensions $s=[s_x,s_y,s_z]$, 
rotation $\theta$ (or `yaw')
of the bounding-box about the gravity-aligned ($y$) axis, 
semantic label $l$ and a feature descriptor vector $\bm{f} \in \mathbb{R}^d$. 
These object attributes are directly extracted from the VoteNet outputs. 
The descriptors are extracted from the PointNet++ \cite{qi2017pointnet++} backbone of VoteNet. For an object proposal, we retrieve the seed points linked to that object and average their PointNet++ faetures.


\begin{figure*}[!htb]
    \centering
    \includegraphics[clip=true, trim= 0in 0in 0in 0in, width=0.95\textwidth]{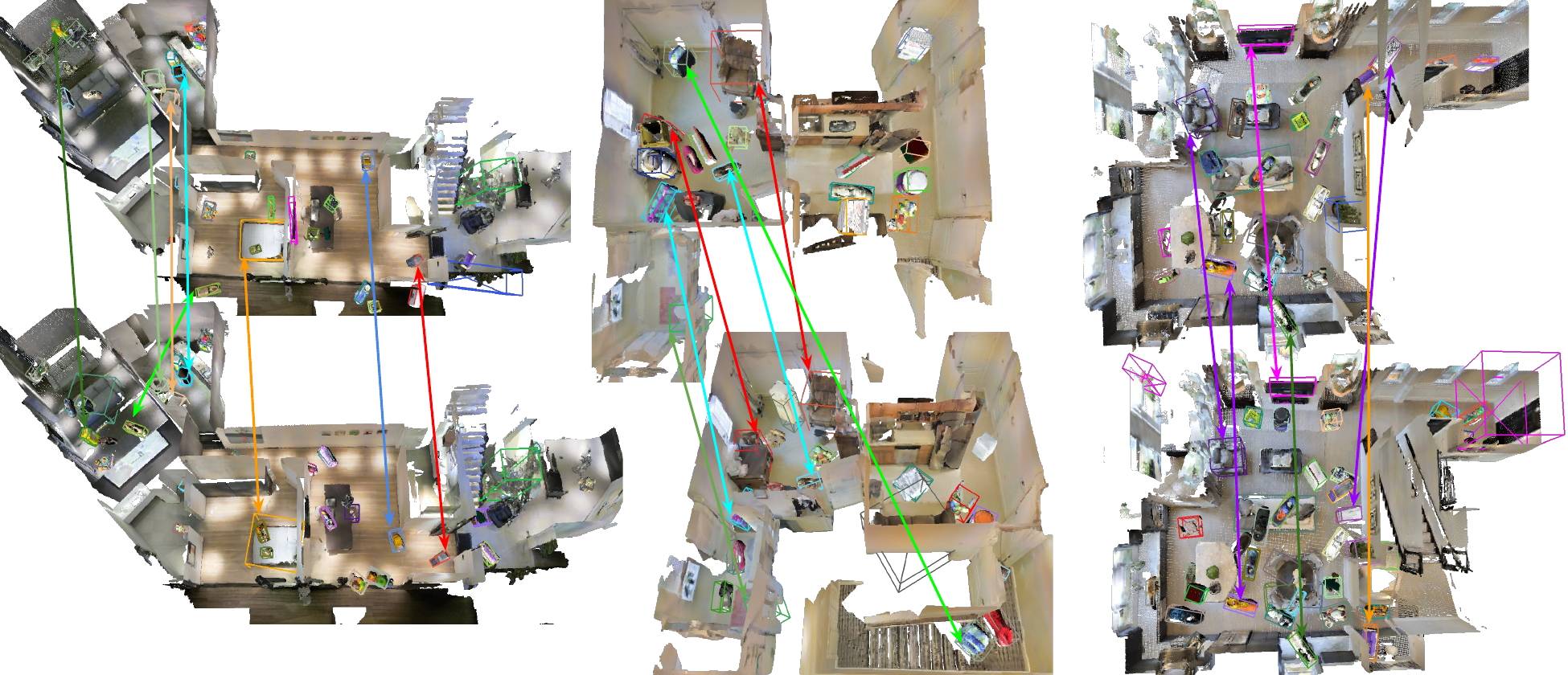}
    \caption{3D-SMNet qualitative results on MP3D scenes \cite{chang2017matterport3d} with inserted YCB \cite{calli2015ycb} and Google-scanned \cite{google_scans} assets. 3D-SMNet is able to match `unchanged` objects like the red examples of the first and second columns. 3D-SMNet is also capable to re-identify objects located at very different locations like the green example of the left column, the green example of the middle column, and the orange example of the right column.}
    \label{fig:3d_smnet_mp3d_results}
\end{figure*}

\subsection{Matcher}
For two layouts $A$ and $B$ with corresponding object-based maps $M_A$ and $M_B$, the goal is to match objects from both maps and identify non-matchable objects (objects added or removed in the scene). We define the two sets of object descriptors $\mathcal{A} = \{\bm{f}^A_1, ..., \bm{f}^A_{m_A}\}$ and $\mathcal{B} = \{\bm{f}^B_1, ..., \bm{f}^B_{m_B}\}$, with $m_A$ and $m_B$ as the total number of objects in $M_A$ and $M_B$. 
The final matching descriptors are computed from a linear projection
$
    \bm{h}^A_i = W  \bm{f}^A_i + \bm{b}, \,\,\, \forall i \in \mathcal{A}, 
$
and similarly for $\mathcal{B}$.

\xhdr{Matching module.} The matching module is based upon the optimal matching layer of SuperGlue \cite{sarlin2020superglue}.  
At a high level, it takes two sets of descriptors and produces an assignment matrix $P$ by solving a linear assignment problem. 
First, we compute the pairwise object scores as the inner product between the descriptors 
$
    S_{i,j} = \langle \bm{h}^A_i, \bm{h}^B_j \rangle, \quad \forall (i,j) \in \mathcal{A} \times \mathcal{B}
$
resulting in a $m_A \times m_B$ score matrix $S$. This matrix can be used to define an assignment problem, with each row $i$ defining the matching scores between object $i$ in $\mathcal{A}$ and all objects in $\mathcal{B}$; and vice versa for column / object $j$ in $\mathcal{B}$.
To account for objects that may not have matches, \cite{sarlin2020superglue} append an additional row and column to $S$ corresponding to `dustbin' objects that objects in $\mathcal{A}$ or $\mathcal{B}$ should match with if they have no true match in the other set. All scores in these dustbin entries are based on a single learnable parameter. 

To be more responsive to individual episodes, we instead compute these dustbin entries based on the object representations. 
Specifically, we add the dustbin column and row as 
$
    S_{i,m_B+1} = z^A_i, S_{m_A+1,j} = z^B_j, 
$
with $\bm{z}^A \in \mathbb{R}^{m_A}$ and $\bm{z}^B \in \mathbb{R}^{m_B}$.
The dustbin score vectors $\bm{z}^A$ and $\bm{z}^B$ are estimated using an attention mechanism from the two sets of matching descriptors $\{h^A_1, ..., h^A_{m_A}\}$ and $\{h^B_1, ..., h^B_{m_B}\}$. We start by computing attention weights from the first set of matching descriptors conditioned on a matching descriptor vector of the other set. The attention weights matrix $W$ is computed from a 3-layer attention MLP module:  
$
    W_{i,j} = \text{MLP}([\bm{h}^A_i; \bm{h}^B_j]), 
$
where $[ ; ]$ is the concatenation operator. To compute the first dustbin score vector $\bm{z}^A$ we softmax the $W$ matrix along the second dimension to obtain the probability matrix 
$
    \hat{W}_{i,j} = \frac{\exp(W_{i,j})}{\sum_{j=1}^{m_B}\exp(W_{i,j})}. 
$
We then compute the attention features as the weighted sum of the matching descriptors:
$
    \bm{a}_i = \sum_{j=1}^{m_B} W_{i,j}  \bm{h}^B_j   \quad \forall i \in \mathcal{A}. 
$
The final values in vector $\bm{z}^A$ are computed from a final 3-layer MLP module:
$
    \bm{z}^A_{i} = \text{MLP}([\bm{h}^A_i; \bm{a}_i]). 
$
The same process is used to compute the dustbin vector $z^B$ by exchanging the roles of $\mathcal{A}$ and $\mathcal{B}$ above.

Given this updated score matrix $S'$ including the dustbin entries, we generate matching assignments using the Sinkhorn algorithm \cite{sinkhorn1967concerning,cuturi2013sinkhorn}, which is a differentiable version of the Hungarian algorithm \cite{munkres1957algorithms} for 
bipartite matching. 
Sinkhorn produces an assignment matrix $P$ where $P_{i,j}$ denotes the probability of matching object $i\in\mathcal{A}$ and $j\in\mathcal{B}$.

\xhdr{Loss.} 3D-SMNet is trained in a supervised fashion using the ground-truth matches $\mathcal{M}=\{ (i,j) \} \in \mathcal{A}\times\mathcal{B}$ and ground-truth unmatchable objects $\mathcal{U^A} \in \mathcal{A}$ and $\mathcal{U^B} \in \mathcal{B}$. Optimization is done by minimizing the negative log-likelihood of the assignment matrix $P$: 
%
\begin{equation}
    L=-\log \prod_{i,j\in \mathcal{M}} P_{ij} \prod_{i \in \mathcal{U^A}} P_{i,m_B+1} \prod_{j \in \mathcal{U^B}} P_{m_A+1,j}
\end{equation}
where the first term guides the model towards matching objects observed in both layouts whereas the latter two terms push unmatchable objects towards the dustbins. As the Sinkhorn algorithm is differentiable, matcher parameters are learned directly to optimize these assignments.


\section{Experiments}

\begin{figure*}[t]
    \centering
    \includegraphics[clip=true, trim=0in 0in 0in 7.1in, width=0.95\textwidth]{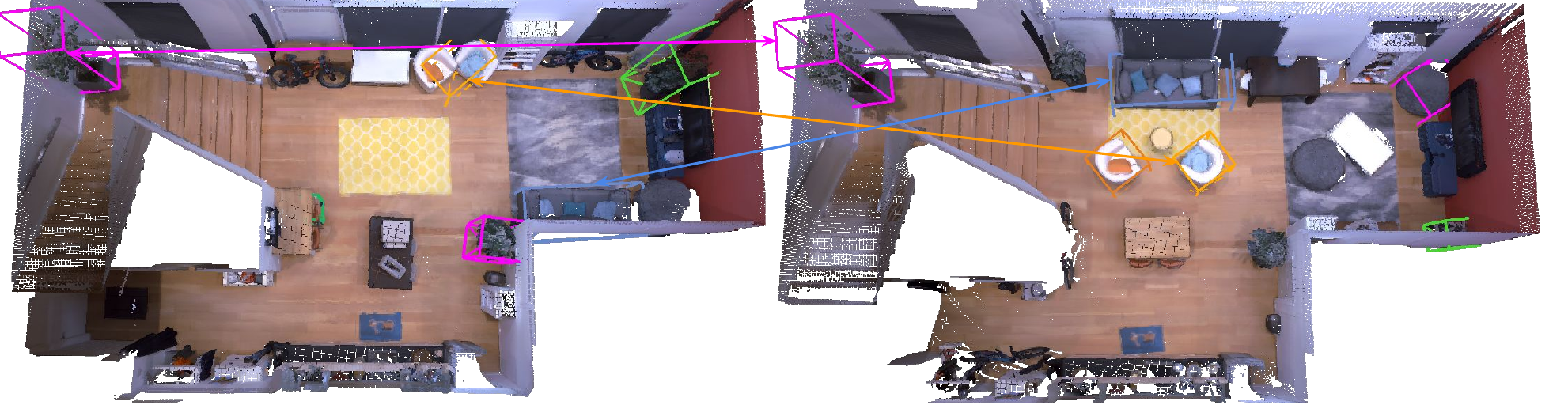}
    \caption{3D-SMNet qualitative results on the Replica scenes \cite{replica19arxiv}. On this zero-shot experiment, 3D-SMNet is able to detect and re-identify `unchanged' objects (like the plant in purple) and `moved' objects (like the couch in blue and chair in orange).}
    \label{fig:3d_smnet_mp3d_results-2}
\vspace{-10pt}
\end{figure*}

\begin{figure}[t]
    \centering
    \includegraphics[clip=true, trim= 0in 0in 0in 0.04in,
    width=1\linewidth]{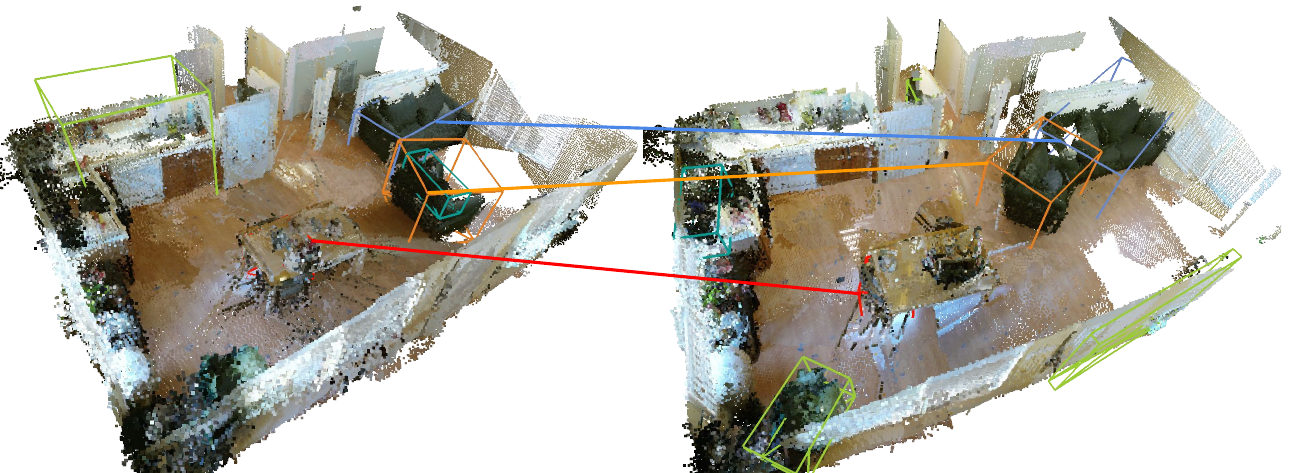}
    \caption{3D-SMNet qualitative results on the Active Vision dataset \cite{replica19arxiv}. On this zero-shot experiment on real data, 3D-SMNet is able to detect and re-identify `unchanged' objects (like the couch in blue) and `moved' objects (like the two chairs in red and chair in orange).}
    \label{fig:3d_smnet_avd_results}
\vspace{-10pt}
\end{figure}

\xhdr{Evaluation Metrics:}
We report the entire range of evaluation metrics for object re-identification in a query-to-gallery setup: Cumulative Matching Characteristics (CMC) \cite{wang2007shape} and mean Average Precision (mAP) \cite{zheng2015scalable}. CMC-k (or rank-k) is the probability  that the correct matched object of a given query object appears in the top-k ranked object list. 
We also report the overall matching accuracy in a multi-query-to-gallery setup (assuming one match per instance possible). It is computed as the total of correctly matched objects in the layout $B$ over the total number of objects in layout $B$. In addition, we measure the overall task performances using accuracy, precision and recall computed from the matches tuples. We start by assigning unique ids to object proposals by matching detections to ground-truth objects using a threshold over 3D-IoUs. From the detections and predicted matches we create object tuples (objA, objB). If a detected object has not been matched, we leave the corresponding entry null in the tuple. A tuple is considered as a true positive (TP) if the same tuple exists in the ground-truth matches. We similarly compute false positive (FP) and false negative (FN), and report the corresponding accuracy, prediction and recall.
These metrics also account for detection performances.

\xhdr{Implementation details:}
We choose a confidence threshold of 0.99 to retain object proposals. The extracted object descriptors are of dimension $256$. The attention MLP module and the final MLP module both have the same architecture with three layers of dimensions $256$, $64$ and $1$. Training takes approximately 6 hours on a single Titan-X with a batch-size of 8 and a learning rate of $5\cdot 10^{-3}$.

\xhdr{Experiments with different matchers:}
We compare 3D-SMNet with different matchers. We experiment with the Hungarian algorithm with different score functions. We tested the $L_2$ and Mahalanobis distances, and learning a 1-layer mapping of the descriptors prior to using the $L_2$ distance to compute the score matrix. We train this one linear layer using a triplet loss function \cite{balntas2016learning}.
\begin{equation}
    L(a,p,n) = \max\{ ||a-p||_2 - ||a-n||_2 + m, 0  \}, 
\end{equation}
with $a,p,n$ the anchor, positive and negative object descriptors. We experimentally set the margin value $m$ to $1.0$ (see suppl. for more details).
In addition, we compare our model to a Sinkhorn matcher (S) without the attention model to estimate the dustbin vectors $\bm{z}^A$ and $\bm{z}^B$. Instead we set the dustbin vectors values to a single trained parameters as in \cite{sarlin2020superglue}.

\xhdr{Oracle experiments:} To set upper bounds on our study, we run two additional experiments. The first (GTmatch) assumes an oracle matcher. Given two sets of detections, this matcher  perfectly matches the detections. In the second (GTbox), we assume ground-truth detections. In this case, we extract features for each ground-truth object using our trained VoteNet \cite{qi2019deep} detector and then match objects using our trained model. Features are extracted by cropping the seed points using the 3D bounding box of the ground-truth object and averaging the seeds' PointNet++ \cite{qi2017pointnet++} features.

\section{Results}
\begin{figure}[t]
    \centering
    \includegraphics[width=0.95\linewidth]{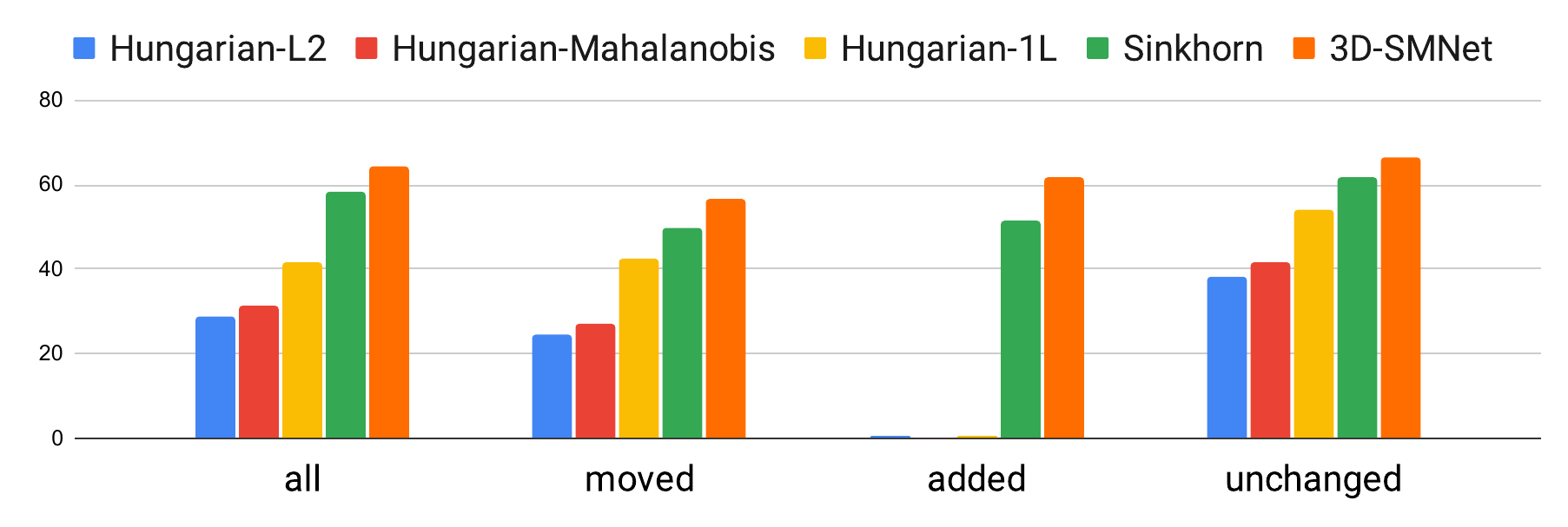}
    \caption{3D-SMNet breakdown matching accuracy per object super-category on the test set of the Matterport scenes \cite{chang2017matterport3d}.}
    \label{fig:3d_smnet_mp3d_results-plot}
\vspace{-10pt}
\end{figure}

Table~\ref{tab:results} shows the matching performances of 3D-SMNet compared to different baselines and 
Fig.~\ref{fig:3d_smnet_mp3d_results-plot} reports the matching metrics of 3D-SMNet split across super-categories (`moved', `added', and `removed'). 

Our first two sets of experiments, Hungarian-L2 (H-L2) and Hungarian-Mahalanobis (H-M) are both working with the Hungarian algorithm solver with the $L_2$  and Mahalanobis distances respectively. We notice that while the Mahalanobis normalized $L_2$ distance increases the matching accuracy slightly ($28.82 \rightarrow 31.44$), for all other metrics (rank@1, rank@5 and mAP) the $L_2$ distance yields better performance (\cref{tab:results} rows 1 and 2). 

\xhdr{Learning a mapping for matching features.} Our third experiment, Hungarian 1L (H-1L), uses the Hungarian algorithm with a $L_2$ distance score. However, prior to measuring the distance, we map the descriptors to a higher dimensional feature  using a linear layer trained under a triplet loss~\cite{balntas2016learning}. This leads to an increase in performance on all metrics (line 3 \vs lines 1 and 2 of \cref{tab:results}) and across all super-categories of objects (yellow bars compared to the blue and red ones from \cref{fig:3d_smnet_mp3d_results-plot}). This suggests that good features for matching can be derived from the extracted descriptors of VoteNet~\cite{qi2019deep}. Additionally, we observe from \cref{fig:3d_smnet_mp3d_results-plot} that on this first set of three experiments (H-L2, H-M, H-1L)  the performance on the `added' objects sup-category is overall very low ($<0.5\%$ matching accuracy). We explain this result by the fact that the Hungarian algorithm cannot, by design, properly handle objects without any counterpart. 

\begin{table*}[t]
\centering
\renewcommand{\arraystretch}{1.15}
\resizebox{\textwidth}{!}{
\begin{tabular}{ l s c c c c s c c c c s c s c c c c }
\toprule
& & \multicolumn{4}{c}{MP3D} &  & \multicolumn{4}{c}{Replica} \\
\cline{3-6}\cline{8-11}
& & rank@1 & rank@5 & mAP & Acc &  & rank@1 & rank@5 & mAP & Acc\\ 
\midrule
H-L2 &  & 42.32 {\scriptsize $\pm$ 0.12 } & 70.94 {\scriptsize $\pm$ 0.13 } & 55.51 {\scriptsize $\pm$ 0.11 } & 28.82 {\scriptsize $\pm$ 0.09 } & &  41.59 {\scriptsize $\pm$ 0.23 } & \textbf{100.00 {\scriptsize $\pm$ 0.00 }} & \textbf{62.54 {\scriptsize $\pm$ 0.15 }} & 24.91 {\scriptsize $\pm$ 0.11 }\\  
H-M &  & 38.18 {\scriptsize $\pm$ 0.13 } & 58.74 {\scriptsize $\pm$ 0.13 } & 48.24 {\scriptsize $\pm$ 0.12 } & 31.44 {\scriptsize $\pm$ 0.10 } & & \textbf{41.83 {\scriptsize $\pm$ 0.20 }} & 95.84 {\scriptsize $\pm$ 0.08 } & 61.21 {\scriptsize $\pm$ 0.13 } & 21.42 {\scriptsize $\pm$ 0.12 } \\  
H-1L &  & 62.18 {\scriptsize $\pm$ 0.10 } & 90.53 {\scriptsize $\pm$ 0.06 } & 74.45 {\scriptsize $\pm$ 0.08 } & 41.70 {\scriptsize $\pm$ 0.07 } & & 33.28 {\scriptsize $\pm$ 0.22 } & 95.77 {\scriptsize $\pm$ 0.08 } & 55.86 {\scriptsize $\pm$ 0.16 } & 25.02 {\scriptsize $\pm$ 0.09 }\\  
Sinkhorn &  & 68.83 {\scriptsize $\pm$ 0.07 } & 94.42 {\scriptsize $\pm$ 0.04 } & 79.75 {\scriptsize $\pm$ 0.06 } & 58.33 {\scriptsize $\pm$ 0.05 } & & 21.12 {\scriptsize $\pm$ 0.14 } & 87.54 {\scriptsize $\pm$ 0.12 } & 47.43 {\scriptsize $\pm$ 0.12 } & 30.51 {\scriptsize $\pm$ 0.10 } \\  
3D-SMNet &  & \textbf{72.85 {\scriptsize $\pm$ 0.08 }} & \textbf{94.84 {\scriptsize $\pm$ 0.04 }} & \textbf{82.36 {\scriptsize $\pm$ 0.06 }} & \textbf{64.35 {\scriptsize $\pm$ 0.06 }} & & 29.29 {\scriptsize $\pm$ 0.14 } & 95.82 {\scriptsize $\pm$ 0.08 } & 53.63 {\scriptsize $\pm$ 0.11 } & \textbf{33.88 {\scriptsize $\pm$ 0.10 }} \\ 
\midrule
GTbox &  & 87.74 {\scriptsize $\pm$ 0.06 } & 98.83 {\scriptsize $\pm$ 0.01 } & 92.49 {\scriptsize $\pm$ 0.04 } & 81.30 {\scriptsize $\pm$ 0.05 } &  & 65.66 {\scriptsize $\pm$ 0.06 } & 97.23 {\scriptsize $\pm$ 0.02 } & 79.92 {\scriptsize $\pm$ 0.04 } & 52.63 {\scriptsize $\pm$ 0.06 } \\
\bottomrule
\end{tabular}}
\caption{3D-SMNet test-set matching results on the Matterport scenes \cite{chang2017matterport3d} with  YCB \cite{calli2015ycb} and Google-scanned \cite{google_scans} assets and on the Replica scenes \cite{replica19arxiv}. 3D-SMNet (line 5) outperforms all the baselines (lines 1-4) on all the metrics on the Matterport scenes. 3D-SMNet performs best on the matching accuracy on the zero-shot experiments with the Replica scenes. The GTbox experiment reports numbers working with ground-truth detections setting up an upper bound for our study.}
\label{tab:results}
\end{table*}

\xhdr{Handling added and removed objects.} 
Recall that our matcher adds an extra dustbin column and row to the score matrix to handle unmatchable objects. 
We see a drastic improvement of the matching accuracy on the `added' objects with $+51\%$ compared to the classical Hungarian baselines (see green bar compared to the blue, red and yellows ones under the `added' category of \cref{fig:3d_smnet_mp3d_results-plot}). 
This also leads to an increase in all metrics on all object super-categories as we see from \cref{tab:results} comparing line 4 to the lines 1-3.

\xhdr{Estimation of the dustbin scores per episode.}
Putting all the pieces together, 3D-SMNet shows the best performances on all metrics across the board (line 5 of \cref{tab:results}) with $+6\%$ on matching accuracy, $+2.6\%$ on mAP, and $+4\%$ on rank@1. We also see a similar increase in performance for all super-categories (orange bar of \cref{fig:3d_smnet_mp3d_results-plot}). This suggests that predicting the dustbin score values given two sets of descriptors is valuable. We experimentally find that across all episodes in the test set the average and standard deviation of the dustbin values are $46.25$ and $29.24$. The large variance in dustbin values highlights the importance of considering multiple values over just a single parameter.
In addition, we also compare 3D-SMNet to an experiment conducted with ground-truth bounding boxes (GTbox). We observe a drop in performance (lines 6 and 5 from \cref{tab:results}) highlighting the impact of detections on matching.
Qualitatively, we find that 3D-SMNet yields good results. \cref{fig:3d_smnet_mp3d_results} reports three samples with detection and re-identification results on  Matterport~\cite{chang2017matterport3d} with new assets~\cite{calli2015ycb,google_scans}. 3D-SMNet is capable of re-identifying objects detected at different locations as depicted by the bright green examples of the first and second columns in \cref{fig:3d_smnet_mp3d_results}, and the orange example in the third column.

\begin{table}[t]
\centering
\renewcommand{\arraystretch}{1.15}
\resizebox{0.95\linewidth}{!}{
\begin{tabular}{l s c c c c }
\toprule
training dataset & & rank@1 & rank@5 & mAP & Acc \\ 
 \midrule
  MP3D &  & 49.46 {\scriptsize $\pm$ 0.50 } & \textbf{100.00 {\scriptsize $\pm$ 0.00 }} & 70.00 {\scriptsize $\pm$ 0.29 } & 51.49 {\scriptsize $\pm$ 0.31 } \\  
RIO &  & 51.68 {\scriptsize $\pm$ 0.49 } & \textbf{100.00 {\scriptsize $\pm$ 0.00 }} & 71.36 {\scriptsize $\pm$ 0.28 } & 51.40 {\scriptsize $\pm$ 0.29 } \\ 
RIO + MP3D &  & \textbf{62.76 {\scriptsize $\pm$ 0.52 }} & 97.38 {\scriptsize $\pm$ 0.14 } & \textbf{76.53 {\scriptsize $\pm$ 0.32 }} & \textbf{61.13 {\scriptsize $\pm$ 0.33 }} \\  
\bottomrule
\end{tabular}}
\caption{3D-SMNet matching results on the validation set of RIO \cite{wald2019rio} when trained with different datasets. 3D-SMNet performs best when trained jointly on real (RIO) and simulated (MP3D) episodes.}
\label{tab:results_rio_dataaug}
\end{table}

\xhdr{Using simulated episodes as data augmentation.} We conduct experiments using the RIO dataset \cite{wald2019rio}. RIO has 1335 point clouds of houses that we select to create 903 episodes split into train and val. We create ground-truth object pairs for each episodes using the instance level annotations of the dataset keeping a subset of categories: \emph{chair, bed, couch, TV, plant and toilet}. We train 3D-SMNet on both the RIO and simulated episodes and compare the performances when the network is trained on RIO or simulated episodes separately. We find from Tab. \ref{tab:results_rio_dataaug} that using simulated episodes as additional data helps with an increase in performances of $+10\%$ on matching accuracy and $rank@1$ and $+5\%$ on $mAP$. Additionally, we also notice that training on simulated data only already provides good results. We observe a decrease of only $0.09\%$ on accuracy and $-2.22\%$ on $rank@1$ compared when training on RIO only.

\xhdr{Zero-shot experiments on photorealistic environments.}
Next, we test our method on the photorealistic Replica dataset~\cite{replica19arxiv}. We select the 6 FRL apartment scenes and create 15 episodes by combining pairwise layouts. We consider the 4 overlapping object categories with our previous study: \emph{chair, bed, couch, and potted plant}. The results are shown in \cref{tab:results}. On this zero-shot experiment we observe that our method performs the best in terms of matching accuracy with +$3.3\%$ increase compared to other baselines (see line 5 \vs 1-4). However, 3D-SMNet is outperformed on all other metrics by H-L2 and H-M. We explain this result because, first the Replica scenes do not have many objects added or removed in the scene and therefore the use of dustbins scores becomes obsolete, and second because the three matchers H-1L, S and 3D-SMNet are trained on Matterport scenes and objects. Nevertheless, we find that 3D-SMNet yields convincing qualitative results as depicted in \cref{fig:3d_smnet_mp3d_results-2}. 3D-SMNet is capable of re-identifying objects at the same positions like the plant in purple. More importantly, 3D-SMNet can re-identify objects located at different locations in the apartment. The chair in orange and couch in blue are both detected and re-identified correctly. 

\xhdr{Zero-shot experiments on more challenging real data.} We also test 3D-SMNet on the Active Vision dataset \cite{ammirato2018active}. Active Vision has 9 unique scenes that were scanned two times under two different layouts using a robot equipped with a Microsoft Kinect device. Each scan consists of a sequence of RGB-D images recorded on a grid map of the scene. At each node on the grid, 12 frames are recorded at 30deg rotational intervals. The camera intrinsics and camera poses were originally estimated from structure-from-motion using COLMAP \cite{schoenberger2016sfm}. We reconstruct the point-clouds by unprojecting pixels using a pinhole camera model as described in Sec. \ref{3d_object_detector}. The Active Vision dataset does not provide 3D annotation information that would allow us to measure the performances of 3D-SMNet. Instead we qualitatively assess the results on this dataset (see Fig. \ref{fig:3d_smnet_avd_results}). We observe that despite having a noisier point-cloud, 3D SMNet is able to detect and re-ID objects. On Fig. \ref{fig:3d_smnet_avd_results} 3D-SMNet detects and re-ID `unchanged` objects like the couch in blue and `moved` objects like the two chairs in red and orange.

\begin{table}[t]
\centering
\renewcommand{\arraystretch}{1.15}
\resizebox{\linewidth}{!}{
\begin{tabular}{ l s c c c s c c c s c c c}
\toprule
& & \multicolumn{3}{c}{MP3D} &  & \multicolumn{3}{c}{Replica} &  & \multicolumn{3}{c}{RIO}\\
\cline{3-5}\cline{7-9}\cline{11-13}
& & Acc & Precision & Recall &   & Acc & Precision & Recall &   & Acc & Precision & Recall \\ 
\midrule
\scriptsize H-L2 &  & 10.29 {\tiny$\pm$0.04 } &21.66 {\tiny$\pm$0.07 } &16.39 {\tiny$\pm$0.05 } &  & 2.63 {\tiny$\pm$0.01 } &6.88 {\tiny$\pm$0.04 } &4.09 {\tiny$\pm$0.02 } &  & 2.82 {\tiny $\pm$ 0.02 } &6.31 {\tiny $\pm$ 0.06 } &4.86 {\tiny $\pm$ 0.04 } \\  

\scriptsize H-M &  & 10.84 {\tiny$\pm$0.03 } &22.70 {\tiny$\pm$0.06 } &17.18 {\tiny$\pm$0.05 } &  & 2.70 {\tiny$\pm$0.02 } &7.08 {\tiny$\pm$0.05 } &4.18 {\tiny$\pm$0.03 } &  & 2.94 {\tiny $\pm$ 0.02 } &6.50 {\tiny $\pm$ 0.03 } &4.95 {\tiny $\pm$ 0.03 } \\  

\scriptsize H-1L &  & 15.19 {\tiny$\pm$0.03 } &30.62 {\tiny$\pm$0.06 } &23.16 {\tiny$\pm$0.04 } &  & 2.68 {\tiny$\pm$0.01 } &7.00 {\tiny$\pm$0.04 } &4.15 {\tiny$\pm$0.02 }  &  & 2.89 {\tiny $\pm$ 0.02 } &6.41 {\tiny $\pm$ 0.05 } &5.00 {\tiny $\pm$ 0.04 } \\  

\scriptsize Sinkhorn &  & 21.08 {\tiny$\pm$0.03 } &39.48 {\tiny$\pm$0.05 } &31.15 {\tiny$\pm$0.04 } &  & 2.82 {\tiny$\pm$0.01 } &7.30 {\tiny$\pm$0.03} &4.40 {\tiny$\pm$0.02 } &  & 3.07 {\tiny $\pm$ 0.03 } &6.78 {\tiny $\pm$ 0.06 } &5.34 {\tiny $\pm$ 0.04 } \\  

\scriptsize 3D-SMNet &  & \textbf{24.59 {\tiny$\pm$0.04 }} & \textbf{44.86 {\tiny$\pm$0.06 }} & \textbf{35.23 {\tiny$\pm$0.05 }} &  & \textbf{3.31 {\tiny$\pm$0.01 }} & \textbf{8.55 {\tiny$\pm$0.04 }} & \textbf{5.12 {\tiny$\pm$0.02 }} &  & \textbf{3.64 {\tiny$\pm$0.03 }} & \textbf{8.01 {\tiny$\pm$0.06 }} & \textbf{6.26 {\tiny$\pm$0.05 }}\\  
\midrule
\scriptsize GTmatch &  & 40.38 {\tiny$\pm$0.05 } &65.09 {\tiny$\pm$0.06 } &51.54 {\tiny$\pm$0.06 }&  & 10.32 {\tiny$\pm$0.03 } &24.41 {\tiny$\pm$0.08 } &15.16{\tiny $\pm$0.04 } &  & 5.77 {\tiny $\pm$ 0.04 } &12.40 {\tiny $\pm$ 0.08 } &9.76 {\tiny $\pm$ 0.07 } \\  
\scriptsize GTbox &  & 69.29 {\tiny$\pm$0.07 } &81.86 {\tiny$\pm$0.05 } &81.86 {\tiny$\pm$0.05 }&  & 29.58 {\tiny$\pm$0.05 } &45.65 {\tiny$\pm$0.06 } &45.65{\tiny $\pm$0.06 } &  & 35.77 {\tiny$\pm$0.18 } &52.64 {\tiny$\pm$0.19 } &52.64 {\tiny$\pm$0.19 } \\
\bottomrule
\end{tabular}}
\caption{3D-SMNet detection and re-ID performances on Matterport \cite{chang2017matterport3d}, Replica \cite{replica19arxiv} and RIO \cite{wald2019rio} scenes. 3D-SMNet (line 5) outperforms the baselines (lines 1-4) on all metrics. The GTbox and GTmatch rows report numbers working with ground-truth detections and an oracle matcher, setting up an upper bound for our experiment.}
\label{tab:results_overall}
\vspace{-10pt}
\end{table}

\xhdr{Detection and re-Identification measured jointly.}
We also measure the overall task performance (detection and re-ID) with accuracy, precision and recall metrics in Table \ref{tab:results_overall}. We find that 3D-SMNet outperforms all baselines on both the Matteport and Replica datasets (rows 5 compared to 1-4). We also report experiments with ground-truth detections (GTbox) and and an oracle matcher (GTmatch). We notice a gap in performance comparing 3D-SMNet with GTmatch and GTbox (lines 5 and 6). This suggests there is room for improvements for detection, $-44.75\%$ drop on Acc from 3D-SMNet to GTBox on Matterport $-26.27\%$ on Replica and $-32.13\%$ on RIO. We make a similar observation on the matcher side with a drop on Acc of $-15.8$ on Matterport, $-7.01$ on Replica and $-2.12\%$ on RIO.

\section{Conclusion}

In this work, we studied the task of object re-identification in 3D environments across egocentric tours. Our proposed object-based map model called 3D-SMNet converts localized egocentric RGB-D videos to sets of localized 3D object bounding boxes with associated feature vectors. We develop an automated pipeline to generate paired egocentric tours of 3D environments under different layouts where objects may be moved, removed, retained, or added. We evaluate our model on this dataset and established a competitive baseline for the task. We also train and evaluate 3D-SMNet on real data. Furthermore, we show jointly training on real and simulated episodes lead to significant improvements over training on real data only. 3D-SMNet also demonstrates great performances on zero-shot experiments on real data. 


{
    \small
    \bibliographystyle{ieeenat_fullname}
    \bibliography{main}
}

\newpage

\section{Appendix}

\subsection{Details on the new assets}

\begin{figure}[h]
    \centering
    \includegraphics[width=0.65\linewidth]{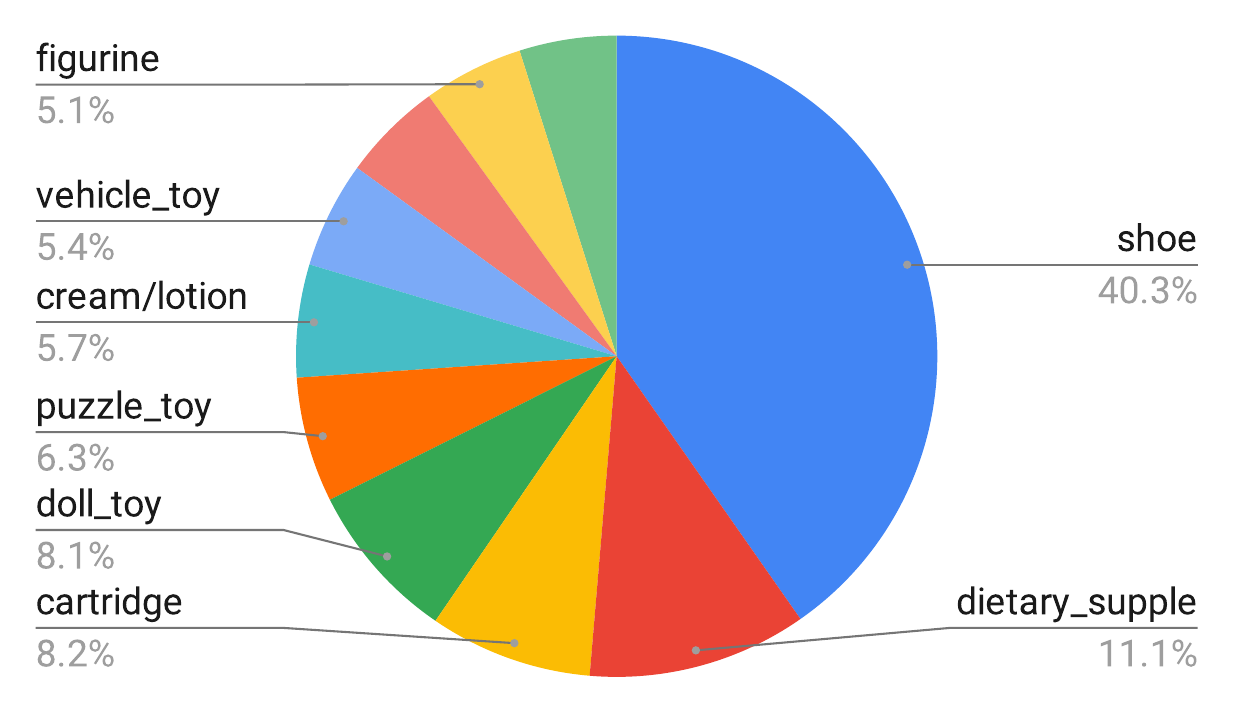}
    \caption{Category distribution of assets inserted in the simulated episodes in Habitat \cite{habitat19iccv} with the Matterport3D scenes \cite{chang2017matterport3d}.}
    \label{fig:all_assets_dist}
\end{figure}

\begin{figure*}[h]
    \centering
    \includegraphics[width=0.95\linewidth]{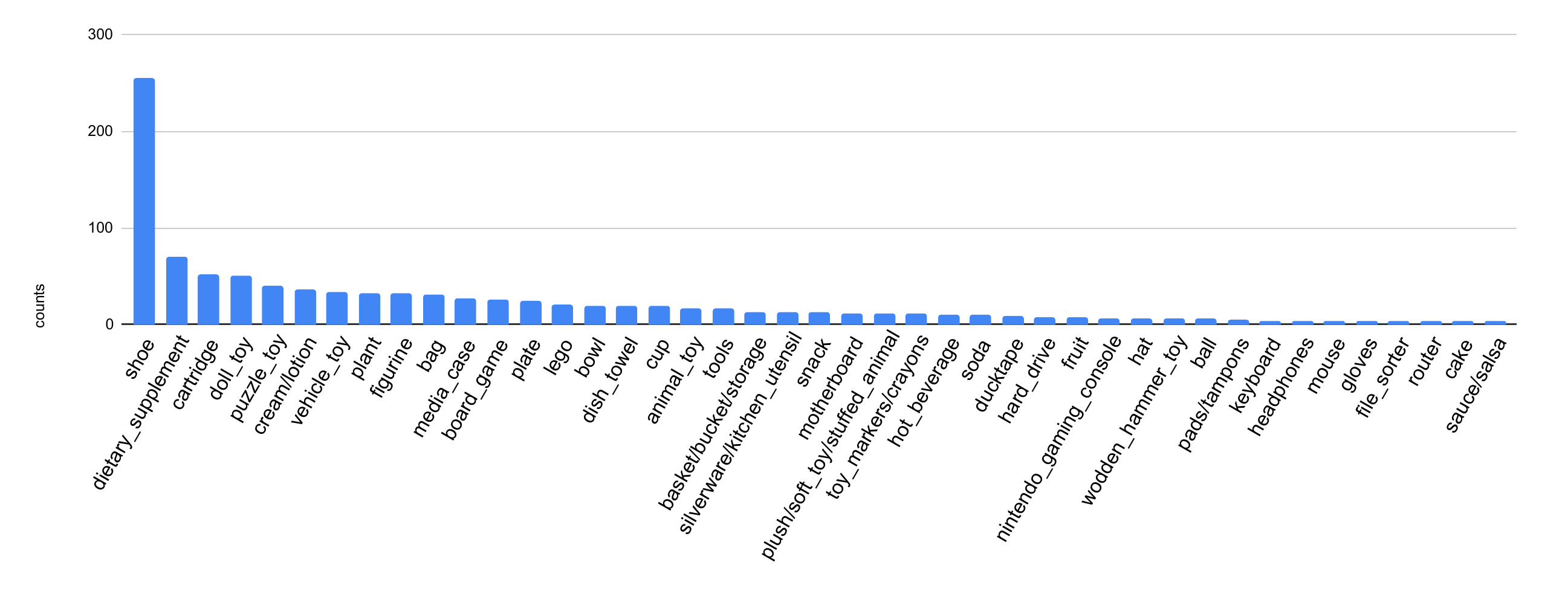}
    \caption{Category distribution of all the Google-scanned objects. We keep the 10 most predominant catgories in our study (up to, and including, the `bag` category)}.
    \label{fig:gso_all_labels}
\end{figure*}

We start by manually labeling assets in the Google-scanned objects dataset \cite{google_scans} that do not have a semantic label. We also re-labeled assets with pre-existing semantic class that are too broad (\eg `consumer goods'). Assets are labeled using names in an unconstrained fashion. We simply assign the name that best describe the object. We then cluster all the labels into 43 object categories shown in \cref{fig:gso_all_labels}. From that list we decide to keep the top 10 classes with the most number of instances and combined them with the YCB dataset \cite{calli2015ycb}. The final object distribution is shown in \cref{fig:all_assets_dist}.

Each object consists of a 3D model. Following this procedure we select 632 models that we split into train/val/test per category. We ensure that all object categories are represented in each split and no model repeats across splits.
In addition, we scale up the models to unrealistic sizes in order to improve their visibility in the egocentric view and to better align with the 3D object detectors capacities. For each object we sample a size variable $t$ from a uniform distribution between $1.0m$ and $1.5m$. We then compute a scaling factor $s$ using the sampled size $t$ and the length $l$ of the largest side of the 3D model as $s = \frac{t}{l}$. Lastly, we scale up the model along the three dimensions using the scaling factor $s$.

\subsection{Example of the generated episodes and trajectories}
\begin{figure*}[t]
    \centering
    \includegraphics[width=\textwidth]{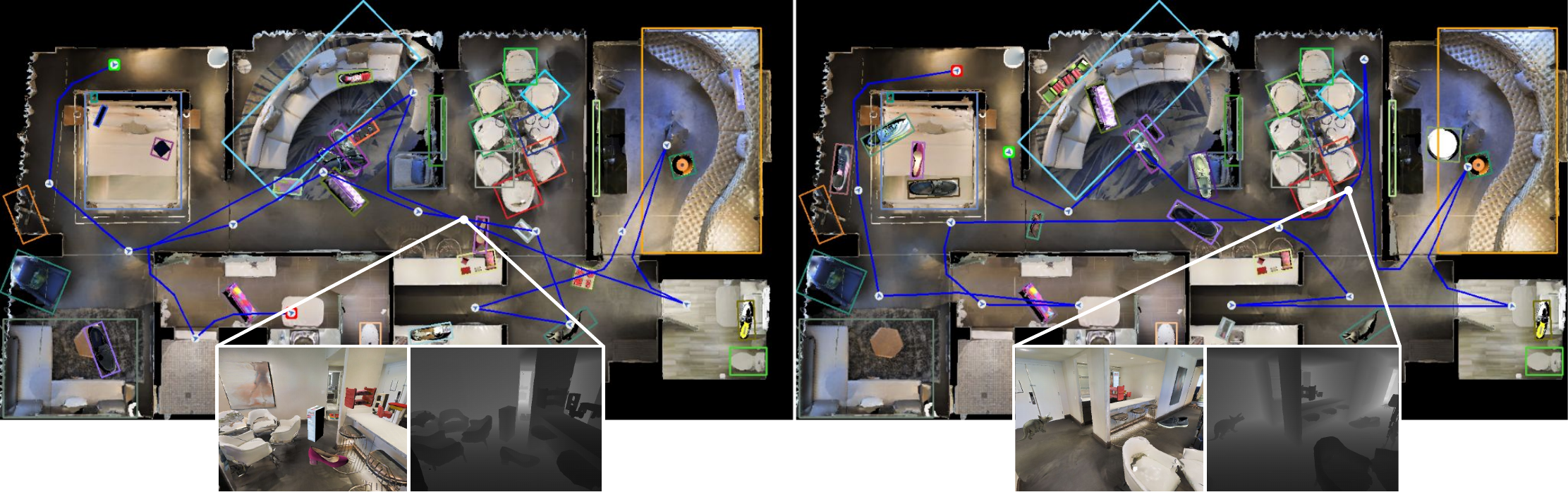}
    \caption{Example multi-object re-identification episode: The two layouts are shown in a top-down view, the agent exploration trajectories are drawn in blue, and bounding boxes show objects of interest. Colors of bounding-boxes indicate correspondences.} 
    \label{fig:3d_smnet_data_sample}
\vspace{-10pt}
\end{figure*}
Recall that the task is to re-identify objects present in both layouts and identify objects which have been
removed or added. Specifically, we do not constrict the two tours to follow the same path. An example episode is shown
in \cref{fig:3d_smnet_data_sample} from a top-down view. Blue lines indicate the tours and objects are marked with bounding boxes. \cref{fig:3d_smnet_data_sample} also shows the downward-looking egocentric view.

\subsection{Details on architectures and feature extraction.}

Each detection is associated with a vector descriptor extracted from the VoteNet backbone: PointNet++ \cite{qi2017pointnet++}. We start by retrieving the cluster of seed points related to that detection. We then average the PointNet++ features of all the seeds in that cluster.
The extracted features from VoteNet \cite{qi2019deep} have a dimension of $256$.
The linear layer used to compute the final matching descriptors has a dimension of $256$. The attention MLP and the final MLP modules have the same architecture with three linear layers of dimensions $256$, $64$ and $1$ combined with rectified linear unit functions.

\subsection{VoteNet \cite{qi2019deep} training details}
\cref{tab:votenet_results} shows the VoteNet performances under different training regimes. We train VoteNet on small crops of the full point cloud. We select $4\times3\times4m$ regions (in the $(x,y,z)$ dimensions) centered around objects of interests.
We conduct multiple experiments on training VoteNet with crops selected from the entire point-cloud using a sliding window approach with a stride of $2m$, $1.5m$ and $2m$ in the $x$, $y$ and $z$ directions. We find that training VoteNet with crops centered around objects leads to the best generalization performances (lines 2-3, lines 4-5 and lines 6-7 from \cref{tab:votenet_results}).
The network is trained from scratch using a batch size of $8$ and a learning rate of $1e^{-3}$. VoteNet is originally designed to work with un-textured point cloud. We extended its architecture to include the RGB information as input and found that it significantly improves the performances (lines 1-2 from \cref{tab:votenet_results}).
In addition, we use the data augmentation technique proposed by VoteNet which consist of random flips and random rotations along the up-right axis. We find that using data augmentation VoteNet performs better (lines 3 and 5 from \cref{tab:votenet_results}). We also use the scheduler from VoteNet and find that it boosts the performances a little more (lines 5 and 7 from \cref{tab:votenet_results}).

\begin{table}[h]
\centering
\caption{VoteNet performances on the validation set tested under different training settings.}
\renewcommand{\arraystretch}{1.15}
\resizebox{\linewidth}{!}{
\begin{tabular}{ c c c c s c c }
\toprule
  object-centered crops & RGB & data aug. & scheduler & & $mAP@25$ & $mAP@50$ \\ 
\midrule
  - & - & - & - &  & 50.566 & 29.963 \\
 \midrule
  - & \checkmark & - & - & & 54.355 & 33.369  \\
  \checkmark & \checkmark & - & - & & 50.582 & 34.265  \\
 \midrule
   - & \checkmark & \checkmark & - & &53.916 & 33.788 \\
  \checkmark & \checkmark & \checkmark & - & & \textbf{58.408} & 38.693  \\
 \midrule
    - & \checkmark & \checkmark & \checkmark & & 56.151 &  35.819 \\
  \checkmark & \checkmark & \checkmark & \checkmark & & 58.170 & \textbf{40.331}  \\
 \midrule
 
\bottomrule
\end{tabular}}

\label{tab:votenet_results}
\end{table}

\subsection{Matching performances when using different features extracted at different stages in VoteNet.}

In this section we study the impact of using different features from VoteNet \cite{qi2019deep} on the final matching results. Specifically, we extract features from the backbone model, PointNet++ \cite{qi2017pointnet++}, and from the last layer of the proposal module of VoteNet. The proposal module is the last element of VoteNet that converts the clustered seed points into object proposal. \cref{tab:matching_results_mp3d_supercat} shows the performance comparison between two sets of experiments using the PointNet++ features (lines 8-13) and the proposal features (lines 1-7). We observe that on all object super-category combined, 3D-SMNet performes best on all metrics when using the PointNet++ features (lines 7 and 12). We also see that the two baselines Sinkhorn and Hungarian-1L have better performances using the PointNet++ features (lines 3 and 10, and lines 6 and 11). The two other baselines Hungarian-L2 and Hungarian-Mahalanobis show better performances when using the proposal features. We explain this phenomena because these two baselines do not involve any training. 

\subsection{Hungarian with one, two or three layers MLP.}
We experiment with a baseline using the Hungarian matching module with the $L_2$ distance to compute the score matrix. Prior to computing the scores we map the descriptors to a matching feature space using a one-linear layer model of dimension $256$. We train this layer using a triplet loss. We create a dataset using the detection proposals of each environment pairs. We start by computing the ground-truth object matches by assigning a unique id to proposals using a 3D-IoU threshold over the set of ground-truth objects.
We create tuples of the form $(a,p,n)$ with an object anchor, its ground-truth match and another object randomly sampled in the scene. $a$, $p$ and $n$ are the anchor, positive and negative object descriptors associated to the object anchor, its match and the other randomly selected object respectively. We set the margin to 1 and use the $L_2$ metric as a measure of distance to train against the triplet loss. We train this model during 50 epoches with a batch size of $64$ and learning rate of $1.5e-2$.
We additionally experiment with a two-layers and three-layers MLP. We found that as we increase the number of layers the performances decrease as shown in lines 3-5 of  \cref{tab:matching_results_mp3d_supercat}, suggesting the model is over-parameterized.

\begin{table*}[h]
\centering
\caption{3D-SMNet test-set matching results on the Matterport scenes \cite{chang2017matterport3d} with new assets detailed over all object super-categories.}
\renewcommand{\arraystretch}{1.15}
\resizebox{\textwidth}{!}{
\begin{tabular}{ l s c s c c c c s c c c c s c s c c c c }
\toprule
& & feature type & & \multicolumn{4}{c}{ALL} &  & \multicolumn{4}{c}{moved} &  & \multicolumn{1}{c}{added} &  & \multicolumn{4}{c}{unchanged}\\
\cline{5-8}\cline{10-13}\cline{15-15}\cline{17-20}
& & & & rank@1 & rank@5 & mAP & Acc &  & rank@1 & rank@5 & mAP & Acc &  & Acc &  & rank@1 & rank@5 & mAP & Acc\\ 
\midrule
 H-L2 & & proposal & & 43.97 {\scriptsize $\pm$ 0.09 } & 69.86 {\scriptsize $\pm$ 0.09 } & 56.06 {\scriptsize $\pm$ 0.09 } & 29.25 {\scriptsize $\pm$ 0.06 } & & 49.28 {\scriptsize $\pm$ 0.16 } & 65.48 {\scriptsize $\pm$ 0.13 } & 57.69 {\scriptsize $\pm$ 0.13 } & 37.93 {\scriptsize $\pm$ 0.13 } & & 1.71 {\scriptsize $\pm$ 0.03 } & & 43.11 {\scriptsize $\pm$ 0.10 } & 70.56 {\scriptsize $\pm$ 0.10 } & 55.80 {\scriptsize $\pm$ 0.10 } & 36.14 {\scriptsize $\pm$ 0.09 } \\ 
 H-M & & proposal & & 45.03 {\scriptsize $\pm$ 0.10 } & 67.97 {\scriptsize $\pm$ 0.11 } & 56.01 {\scriptsize $\pm$ 0.10 } & 35.08 {\scriptsize $\pm$ 0.07 } & & 56.06 {\scriptsize $\pm$ 0.15 } & 75.72 {\scriptsize $\pm$ 0.13 } & 65.82 {\scriptsize $\pm$ 0.13 } & 53.56 {\scriptsize $\pm$ 0.14 } & & 1.15 {\scriptsize $\pm$ 0.03 } & & 43.27 {\scriptsize $\pm$ 0.11 } & 66.74 {\scriptsize $\pm$ 0.12 } & 54.44 {\scriptsize $\pm$ 0.10 } & 42.28 {\scriptsize $\pm$ 0.11 } \\  
 H-1L & & proposal & & 61.64 {\scriptsize $\pm$ 0.09 } & 91.06 {\scriptsize $\pm$ 0.07 } & 74.31 {\scriptsize $\pm$ 0.08 } & 40.75 {\scriptsize $\pm$ 0.06 } & & 74.02 {\scriptsize $\pm$ 0.11 } & 97.15 {\scriptsize $\pm$ 0.04 } & 83.88 {\scriptsize $\pm$ 0.07 } & 61.85 {\scriptsize $\pm$ 0.13 } & & 1.47 {\scriptsize $\pm$ 0.03 } & & 59.67 {\scriptsize $\pm$ 0.10 } & 90.09 {\scriptsize $\pm$ 0.08 } & 72.78 {\scriptsize $\pm$ 0.08 } & 49.14 {\scriptsize $\pm$ 0.10 } \\  
 H-2L & & proposal & & 60.57 {\scriptsize $\pm$ 0.10 } & 91.25 {\scriptsize $\pm$ 0.07 } & 73.81 {\scriptsize $\pm$ 0.08 } & 39.85 {\scriptsize $\pm$ 0.06 } & & 73.34 {\scriptsize $\pm$ 0.14 } & 97.59 {\scriptsize $\pm$ 0.04 } & 83.83 {\scriptsize $\pm$ 0.08 } & 59.59 {\scriptsize $\pm$ 0.12 } & & 1.49 {\scriptsize $\pm$ 0.03 } & & 58.54 {\scriptsize $\pm$ 0.10 } & 90.24 {\scriptsize $\pm$ 0.08 } & 72.22 {\scriptsize $\pm$ 0.09 } & 48.18 {\scriptsize $\pm$ 0.09 } \\  
 H-3L & & proposal & & 59.99 {\scriptsize $\pm$ 0.09 } & 91.38 {\scriptsize $\pm$ 0.07 } & 73.26 {\scriptsize $\pm$ 0.07 } & 38.22 {\scriptsize $\pm$ 0.06 } & & 74.50 {\scriptsize $\pm$ 0.10 } & \textbf{98.47 {\scriptsize $\pm$ 0.04 }} & 84.43 {\scriptsize $\pm$ 0.07 } & 58.06 {\scriptsize $\pm$ 0.13 } & & 1.23 {\scriptsize $\pm$ 0.02 } & & 57.67 {\scriptsize $\pm$ 0.10 } & 90.24 {\scriptsize $\pm$ 0.08 } & 71.47 {\scriptsize $\pm$ 0.08 } & 46.14 {\scriptsize $\pm$ 0.10 } \\ 
 Sinkhorn & & proposal & & 61.44 {\scriptsize $\pm$ 0.10 } & 92.02 {\scriptsize $\pm$ 0.08 } & 74.19 {\scriptsize $\pm$ 0.09 } & 54.85 {\scriptsize $\pm$ 0.06 } & & 76.00 {\scriptsize $\pm$ 0.11 } & 97.64 {\scriptsize $\pm$ 0.04 } & 84.81 {\scriptsize $\pm$ 0.07 } & 66.19 {\scriptsize $\pm$ 0.11 } & & 55.63 {\scriptsize $\pm$ 0.09 } & & 59.11 {\scriptsize $\pm$ 0.11 } & 91.12 {\scriptsize $\pm$ 0.09 } & 72.49 {\scriptsize $\pm$ 0.10 } & 52.78 {\scriptsize $\pm$ 0.09 } \\  
 3D-SMNet & & proposal & & 66.88 {\scriptsize $\pm$ 0.10 } & 92.85 {\scriptsize $\pm$ 0.07 } & 78.19 {\scriptsize $\pm$ 0.08 } & 60.06 {\scriptsize $\pm$ 0.07 } & & \textbf{79.48 {\scriptsize $\pm$ 0.12 }} & 97.91 {\scriptsize $\pm$ 0.04 } & \textbf{87.51 {\scriptsize $\pm$ 0.08 }} & \textbf{68.68 {\scriptsize $\pm$ 0.14 }} & & \textbf{63.10 {\scriptsize $\pm$ 0.09 }} & & 64.88 {\scriptsize $\pm$ 0.11 } & 92.05 {\scriptsize $\pm$ 0.07 } & 76.71 {\scriptsize $\pm$ 0.08 } & 57.76 {\scriptsize $\pm$ 0.10 } \\

\midrule
\midrule
 H-L2 &  & pointnet++ & & 42.15 {\scriptsize $\pm$ 0.12 } & 70.79 {\scriptsize $\pm$ 0.12 } & 55.36 {\scriptsize $\pm$ 0.11 } & 28.69 {\scriptsize $\pm$ 0.09 } & & 27.45 {\scriptsize $\pm$ 0.13 } & 54.12 {\scriptsize $\pm$ 0.16 } & 40.34 {\scriptsize $\pm$ 0.12 } & 24.35 {\scriptsize $\pm$ 0.13 } & & 0.44 {\scriptsize $\pm$ 0.01 } & & 44.41 {\scriptsize $\pm$ 0.14 } & 73.35 {\scriptsize $\pm$ 0.14 } & 57.67 {\scriptsize $\pm$ 0.13 } & 37.96 {\scriptsize $\pm$ 0.13 } \\  
 H-M &  & pointnet++ & & 38.21 {\scriptsize $\pm$ 0.13 } & 58.66 {\scriptsize $\pm$ 0.12 } & 48.23 {\scriptsize $\pm$ 0.12 } & 31.40 {\scriptsize $\pm$ 0.10 } & & 26.61 {\scriptsize $\pm$ 0.12 } & 51.02 {\scriptsize $\pm$ 0.14 } & 37.87 {\scriptsize $\pm$ 0.10 } & 27.45 {\scriptsize $\pm$ 0.13 } & & 0.22 {\scriptsize $\pm$ 0.01 } & & 40.00 {\scriptsize $\pm$ 0.15 } & 59.84 {\scriptsize $\pm$ 0.14 } & 49.82 {\scriptsize $\pm$ 0.14 } & 41.50 {\scriptsize $\pm$ 0.14 } \\  
 H-1L &  & pointnet++ & &62.06 {\scriptsize $\pm$ 0.10 } & 90.52 {\scriptsize $\pm$ 0.06 } & 74.37 {\scriptsize $\pm$ 0.08 } & 41.65 {\scriptsize $\pm$ 0.07 } & & 55.06 {\scriptsize $\pm$ 0.14 } & 84.63 {\scriptsize $\pm$ 0.09 } & 67.94 {\scriptsize $\pm$ 0.11 } & 42.41 {\scriptsize $\pm$ 0.14 } & & 0.45 {\scriptsize $\pm$ 0.02 } & & 63.13 {\scriptsize $\pm$ 0.12 } & 91.43 {\scriptsize $\pm$ 0.07 } & 75.35 {\scriptsize $\pm$ 0.09 } & 54.01 {\scriptsize $\pm$ 0.11 } \\  
 Sinkhorn &  & pointnet++ & & 68.83 {\scriptsize $\pm$ 0.09 } & 94.44 {\scriptsize $\pm$ 0.04 } & 79.77 {\scriptsize $\pm$ 0.07 } & 58.35 {\scriptsize $\pm$ 0.06 } & & 56.91 {\scriptsize $\pm$ 0.14 } & 89.62 {\scriptsize $\pm$ 0.09 } & 70.52 {\scriptsize $\pm$ 0.10 } & 49.88 {\scriptsize $\pm$ 0.11 } & & 51.49 {\scriptsize $\pm$ 0.09 } & & 70.66 {\scriptsize $\pm$ 0.10 } & 95.19 {\scriptsize $\pm$ 0.05 } & 81.19 {\scriptsize $\pm$ 0.08 } & 61.79 {\scriptsize $\pm$ 0.09 } \\  
 3D-SMNet &  &pointnet++ & & \textbf{72.84 {\scriptsize $\pm$ 0.09 }} & \textbf{94.83 {\scriptsize $\pm$ 0.04 }} & \textbf{82.35 {\scriptsize $\pm$ 0.07 }} & \textbf{64.33 {\scriptsize $\pm$ 0.06 }} & & 64.28 {\scriptsize $\pm$ 0.13 } & 92.22 {\scriptsize $\pm$ 0.07 } & 75.98 {\scriptsize $\pm$ 0.09 } & 56.31 {\scriptsize $\pm$ 0.13 } & & 61.81 {\scriptsize $\pm$ 0.08 } & & \textbf{74.16 {\scriptsize $\pm$ 0.10 }} & \textbf{95.23 {\scriptsize $\pm$ 0.05 }} & \textbf{83.33 {\scriptsize $\pm$ 0.08 }} & \textbf{66.37 {\scriptsize $\pm$ 0.09 }} \\  
 \midrule
 GTbox &  &pointnet++ & & 87.74 {\scriptsize $\pm$ 0.06 } & 98.83 {\scriptsize $\pm$ 0.01 } & 92.49 {\scriptsize $\pm$ 0.04 } & 81.27 {\scriptsize $\pm$ 0.05 } & & 85.55 {\scriptsize $\pm$ 0.09 } & 98.58 {\scriptsize $\pm$ 0.03 } & 91.48 {\scriptsize $\pm$ 0.05 } & 78.19 {\scriptsize $\pm$ 0.09 } & & 69.39 {\scriptsize $\pm$ 0.06 } & & 88.02 {\scriptsize $\pm$ 0.06 } & 98.87 {\scriptsize $\pm$ 0.01 } & 92.62 {\scriptsize $\pm$ 0.04 } & 84.49 {\scriptsize $\pm$ 0.06 } \\

\bottomrule
\end{tabular}}

\label{tab:matching_results_mp3d_supercat}
\end{table*}

\begin{table*}[h]
\centering
\caption{3D-SMNet test-set detection and re-ID results on the Matterport scenes \cite{chang2017matterport3d} with new assets detailed over all object super-categories.}
\renewcommand{\arraystretch}{1.15}
\resizebox{\linewidth}{!}{
\begin{tabular}{ l s c c c s c c c s c c c s c c c}
\toprule
& & \multicolumn{3}{c}{All} &  & \multicolumn{3}{c}{moved} &  & \multicolumn{3}{c}{added}&  & \multicolumn{3}{c}{unchanged}\\
\cline{3-5}\cline{7-9}\cline{11-13}\cline{15-17}
& & Acc & Precision & Recall &   & Acc & Precision & Recall &   & Acc & Precision & Recall &   & Acc & Precision & Recall \\ 
\midrule
 H-L2 &  & 10.33 {\scriptsize $\pm$ 0.04 } &21.73 {\scriptsize $\pm$ 0.07 } &16.44 {\scriptsize $\pm$ 0.05 } & & 0.13 {\scriptsize $\pm$ 0.01 } &0.17 {\scriptsize $\pm$ 0.01 } &0.61 {\scriptsize $\pm$ 0.03 } & & 0.45 {\scriptsize $\pm$ 0.01 } &0.58 {\scriptsize $\pm$ 0.02 } &1.87 {\scriptsize $\pm$ 0.06 } & & 10.65 {\scriptsize $\pm$ 0.04 } &24.05 {\scriptsize $\pm$ 0.09 } &16.04 {\scriptsize $\pm$ 0.06 } \\  
H-M &  & 10.88 {\scriptsize $\pm$ 0.04 } &22.77 {\scriptsize $\pm$ 0.07 } &17.24 {\scriptsize $\pm$ 0.05 } & & 0.25 {\scriptsize $\pm$ 0.01 } &0.31 {\scriptsize $\pm$ 0.01 } &1.21 {\scriptsize $\pm$ 0.05 } & & 0.22 {\scriptsize $\pm$ 0.01 } &0.29 {\scriptsize $\pm$ 0.01 } &0.87 {\scriptsize $\pm$ 0.03 } & & 11.09 {\scriptsize $\pm$ 0.04 } &24.91 {\scriptsize $\pm$ 0.09 } &16.66 {\scriptsize $\pm$ 0.06 } \\  
Hungarian-1L &  & 15.24 {\scriptsize $\pm$ 0.03 } &30.71 {\scriptsize $\pm$ 0.06 } &23.23 {\scriptsize $\pm$ 0.04 } & & 0.25 {\scriptsize $\pm$ 0.01 } &0.31 {\scriptsize $\pm$ 0.01 } &1.23 {\scriptsize $\pm$ 0.05 } & & 0.45 {\scriptsize $\pm$ 0.01 } &0.60 {\scriptsize $\pm$ 0.02 } &1.81 {\scriptsize $\pm$ 0.06 } & & 15.70 {\scriptsize $\pm$ 0.04 } &33.89 {\scriptsize $\pm$ 0.07 } &22.62 {\scriptsize $\pm$ 0.05 } \\  
Sinkhorn &  & 21.03 {\scriptsize $\pm$ 0.04 } &39.40 {\scriptsize $\pm$ 0.06 } &31.08 {\scriptsize $\pm$ 0.05 } & & 2.08 {\scriptsize $\pm$ 0.02 } &2.81 {\scriptsize $\pm$ 0.03 } &7.46 {\scriptsize $\pm$ 0.08 } & & 34.19 {\scriptsize $\pm$ 0.09 } &41.04 {\scriptsize $\pm$ 0.11 } &67.22 {\scriptsize $\pm$ 0.11 } & & 17.56 {\scriptsize $\pm$ 0.03 } &36.22 {\scriptsize $\pm$ 0.07 } &25.42 {\scriptsize $\pm$ 0.04 } \\  
3D-SMNet &  & \textbf{24.53 {\scriptsize $\pm$ 0.04 }} & \textbf{44.77 {\scriptsize $\pm$ 0.06 }} & \textbf{35.17 {\scriptsize $\pm$ 0.05 }} & & \textbf{2.25 {\scriptsize $\pm$ 0.03 }} & \textbf{3.06 {\scriptsize $\pm$ 0.04 }} & \textbf{7.87 {\scriptsize $\pm$ 0.10 }} & & \textbf{37.86 {\scriptsize $\pm$ 0.09 }} & \textbf{44.86 {\scriptsize $\pm$ 0.11 }} & \textbf{70.84 {\scriptsize $\pm$ 0.11 }} & & \textbf{20.48 {\scriptsize $\pm$ 0.04 }} & \textbf{41.57 {\scriptsize $\pm$ 0.07 }} & \textbf{28.77 {\scriptsize $\pm$ 0.05 }} \\  
\midrule
GTmatch &  & 40.41 {\scriptsize $\pm$ 0.05 } &65.21 {\scriptsize $\pm$ 0.06 } &51.52 {\scriptsize $\pm$ 0.05 } & & 4.88 {\scriptsize $\pm$ 0.04 } &6.23 {\scriptsize $\pm$ 0.06 } &18.40 {\scriptsize $\pm$ 0.15 } & & 50.97 {\scriptsize $\pm$ 0.09 } &58.34 {\scriptsize $\pm$ 0.10 } &80.15 {\scriptsize $\pm$ 0.08 } & & 34.85 {\scriptsize $\pm$ 0.05 } &63.33 {\scriptsize $\pm$ 0.05 } &43.66 {\scriptsize $\pm$ 0.06 } \\  
GTbox &  & 69.30 {\scriptsize $\pm$ 0.08 } &81.86 {\scriptsize $\pm$ 0.05 } &81.86 {\scriptsize $\pm$ 0.05 } & & 30.18 {\scriptsize $\pm$ 0.25 } &46.25 {\scriptsize $\pm$ 0.29 } &46.25 {\scriptsize $\pm$ 0.29 } & & 100.00 {\scriptsize $\pm$ 0.00 } &100.00 {\scriptsize $\pm$ 0.00 } &100.00 {\scriptsize $\pm$ 0.00 } & & 67.29 {\scriptsize $\pm$ 0.10 } &80.44 {\scriptsize $\pm$ 0.07 } &80.44 {\scriptsize $\pm$ 0.07 } \\ 
\bottomrule
\end{tabular}}
\label{tab:results_overall_mp3d}
\end{table*}

\subsection{Evaluation for each object super-categories.}
\begin{figure*}[!htb]
    \centering
    \includegraphics[width=\textwidth]{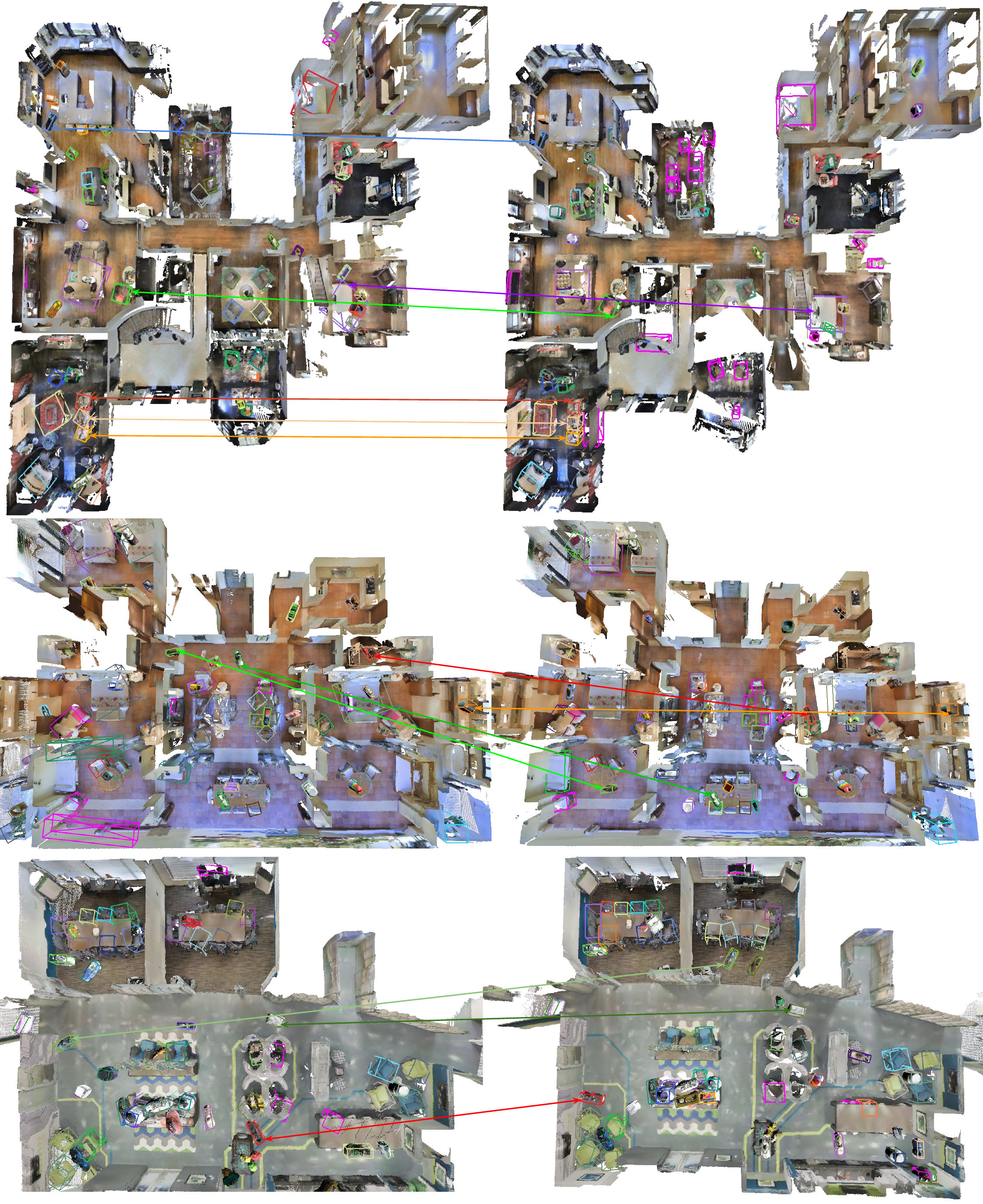}
    \caption{3D-SMNet qualitative results on MP3D scenes \cite{chang2017matterport3d} with inserted YCB \cite{calli2015ycb} and Google-scanned \cite{google_scans} assets.}
    \label{fig:3d_smnet_mp3d_results_supp}
\end{figure*}

\begin{figure*}[!htb]
    \centering
    \includegraphics[width=\textwidth]{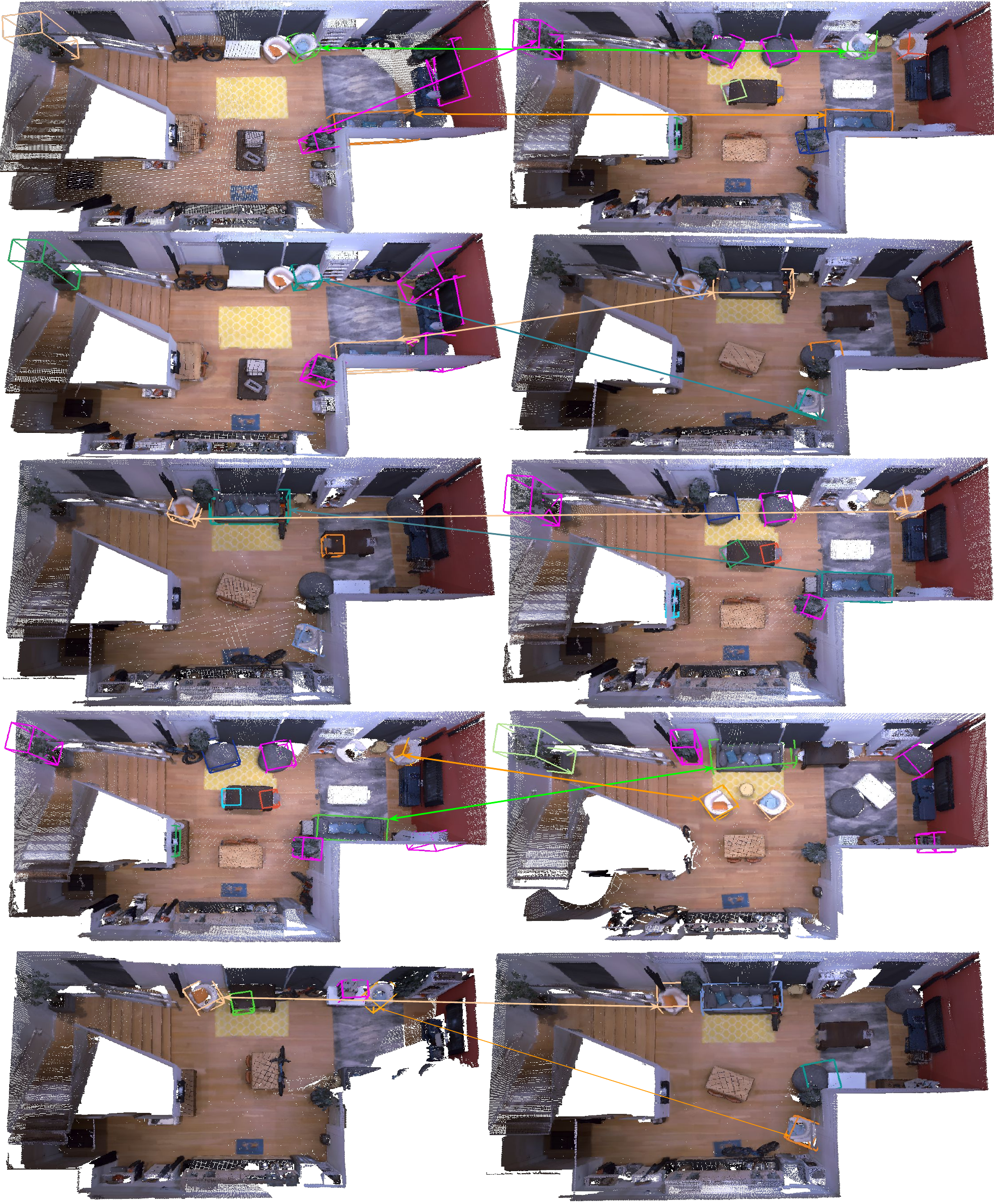}
    \caption{3D-SMNet qualitative results on Replica scenes \cite{replica19arxiv}.}
    \label{fig:3d_smnet_replica_results_supp}
\end{figure*}

\begin{figure*}[!htb]
    \centering
    \includegraphics[width=\textwidth]{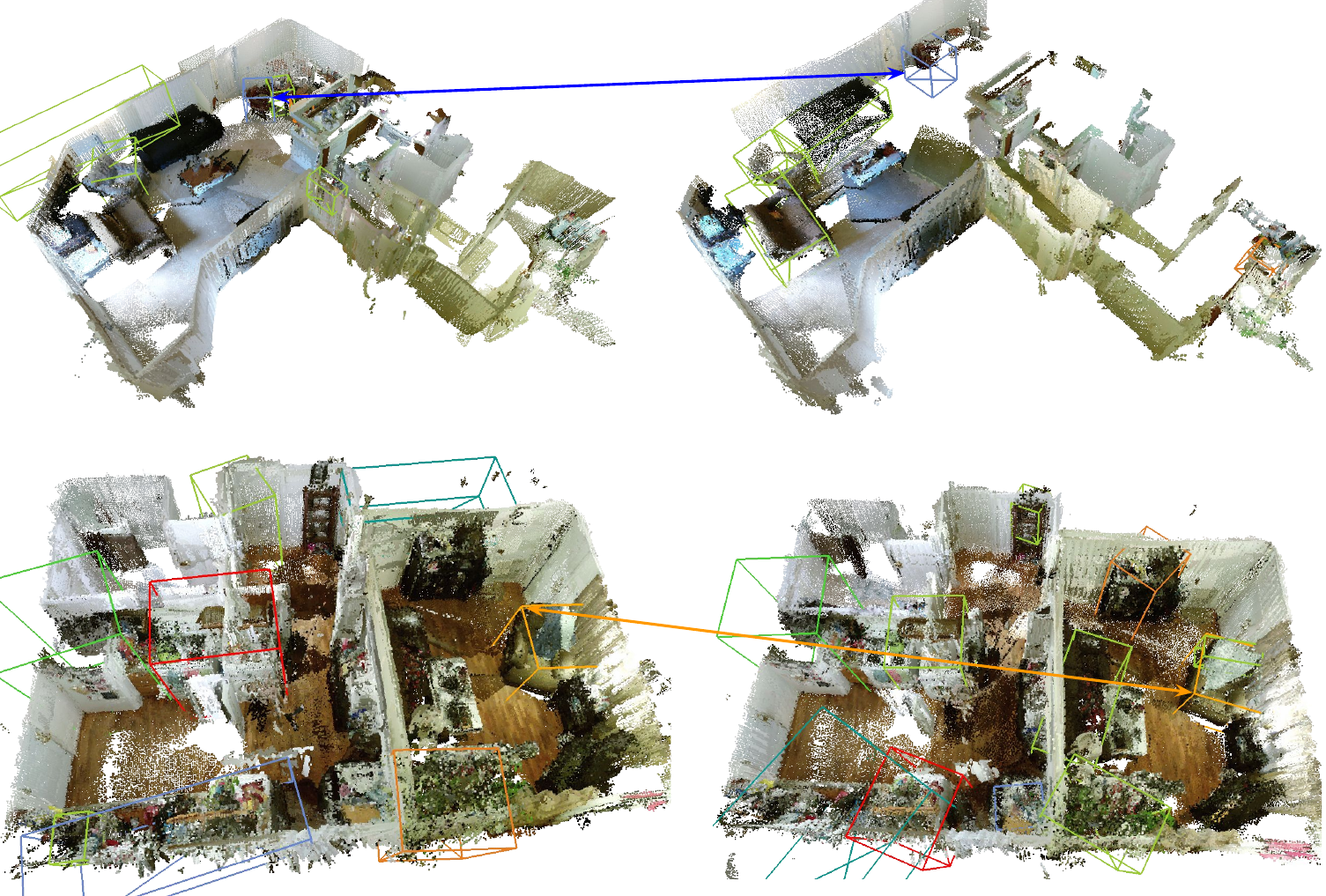}
    \caption{3D-SMNet qualitative results on Active-Vision scenes \cite{ammirato2018active}}
    \label{fig:3d_smnet_avd_results_supp}
\end{figure*}

In \cref{tab:matching_results_mp3d_supercat} we perform matching evaluation for each object super-categories (`added', `moved', `unchanged') in the test set of the Matterport scenes \cite{chang2017matterport3d}. 3D-SMNet consistently outperforms baseline approaches on all super-categories across all metrics (lines 8-12).
\cref{tab:matching_results_mp3d_supercat} also reports numbers on the oracle experiment working with ground-truth detections (GTbox). In this experiment we use the 3D-SMNet matcher module to match ground-truth detections. It measures the impact of the false-positive and false-negative detections on the overall matching performances. We observed a decrease of $17\%$ on the matching accuracy (lines 12 and 13) suggesting that a better detector would lead to better matching outcomes. 
In \cref{tab:results_overall_mp3d} we perform join detection and re-ID evaluation for each object super-categories (`added', `moved', `unchanged') in the test set of the Matterport scenes \cite{chang2017matterport3d}. 3D-SMNet consistently outperforms baseline approaches on all super-categories across all metrics (lines 1-5). \cref{fig:3d_smnet_mp3d_results_supp}, \cref{fig:3d_smnet_replica_results_supp} and \cref{fig:3d_smnet_avd_results_supp} show additional qualitative results on the Matterport \cite{chang2017matterport3d} Replica \cite{replica19arxiv} and Active-Vision \cite{ammirato2018active} scenes.


\subsection{Precision Recall analysis on Replica and RIO.}
\begin{figure*}[h]
    \centering
    \includegraphics[width=0.95\linewidth]{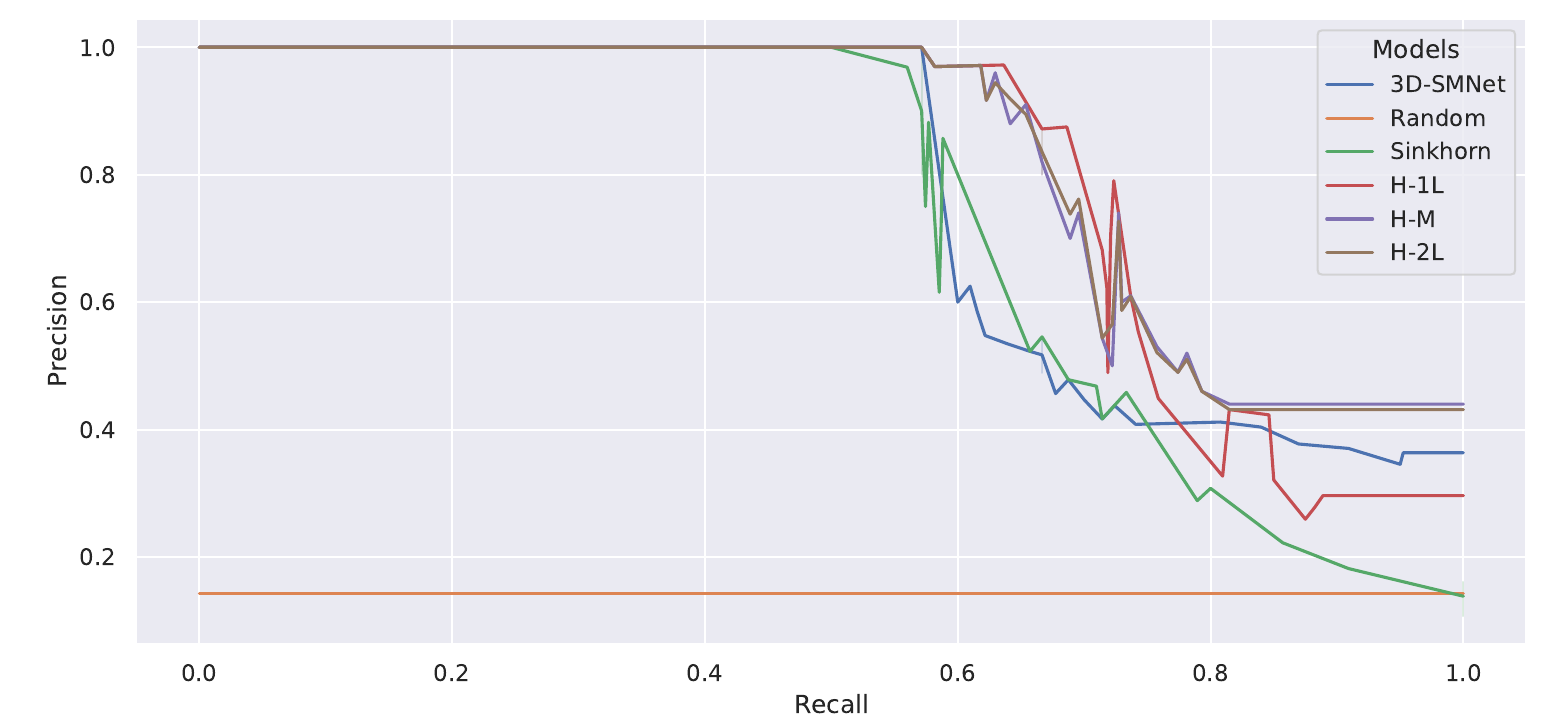}
    \caption{Matching performances precision-recall curves of the different models on the Replica dataset \cite{replica19arxiv}}
    \label{fig:precision_recall_replica}
\end{figure*}

\begin{figure*}[h]
    \centering
    \includegraphics[width=0.95\linewidth]{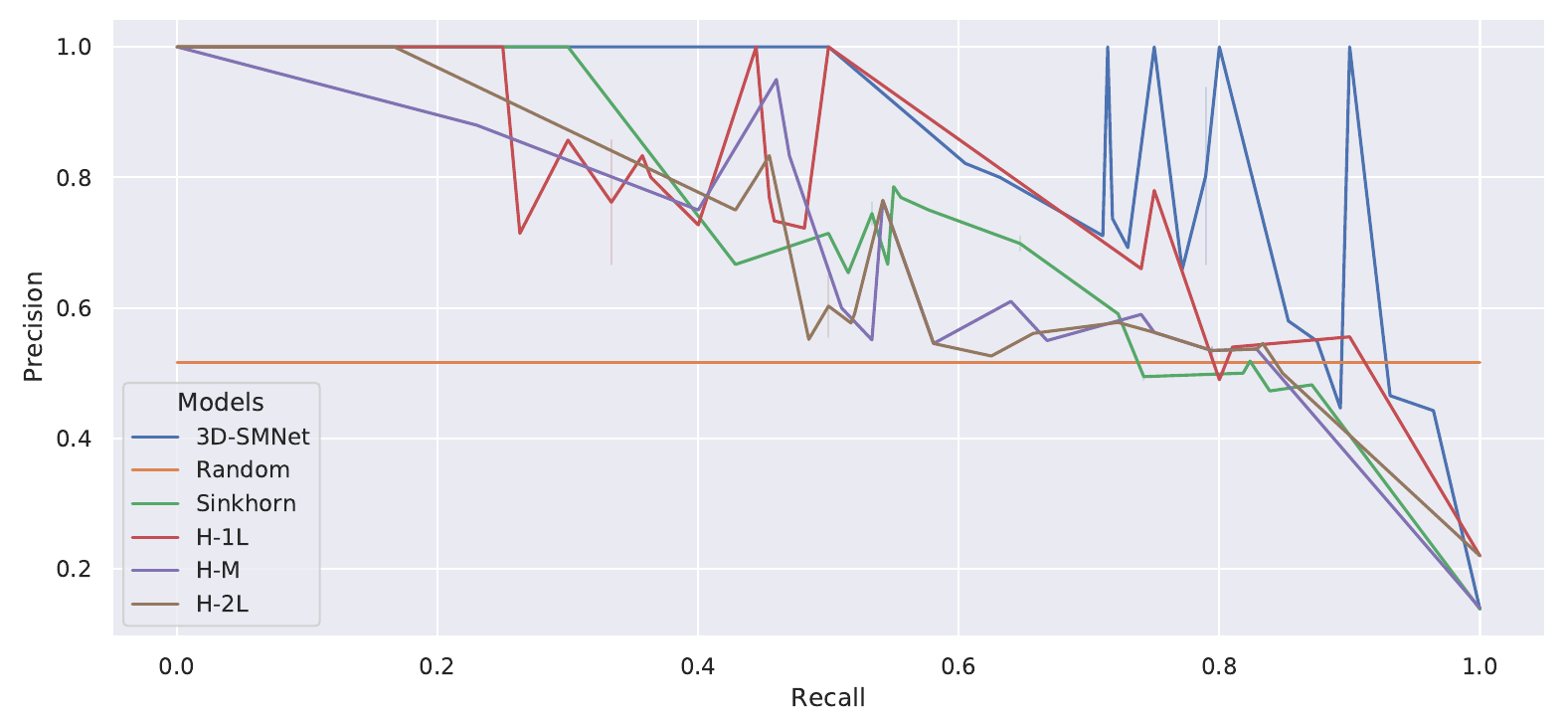}
    \caption{Matching performances precision-recall curves of the different models on the Rio dataset \cite{wald2019rio}}
    \label{fig:precision_recall_rio}
\end{figure*}

We compute the precision-recall curves by varying the matching threshold for the different models on the RIO \cite{wald2019rio} and Replica \cite{replica19arxiv} datasets. We observe on Fig. \ref{fig:precision_recall_replica} that all models perform better than random accross all matching thresholds on the zero-shot transfer on Replica scenes. We notice on Fig. \ref{fig:precision_recall_rio} that above a certain matching threshold models tends to perform worst than random on the RIO scenes. The precision value for the random model is higher due to a smaller number of object instances present in scenes. Nevertheless, 3D-SMNet performs better with a precision of $1.0$ at recall $0.9$. 

\begin{figure*}[!htb]
    \centering
    \includegraphics[width=\textwidth]{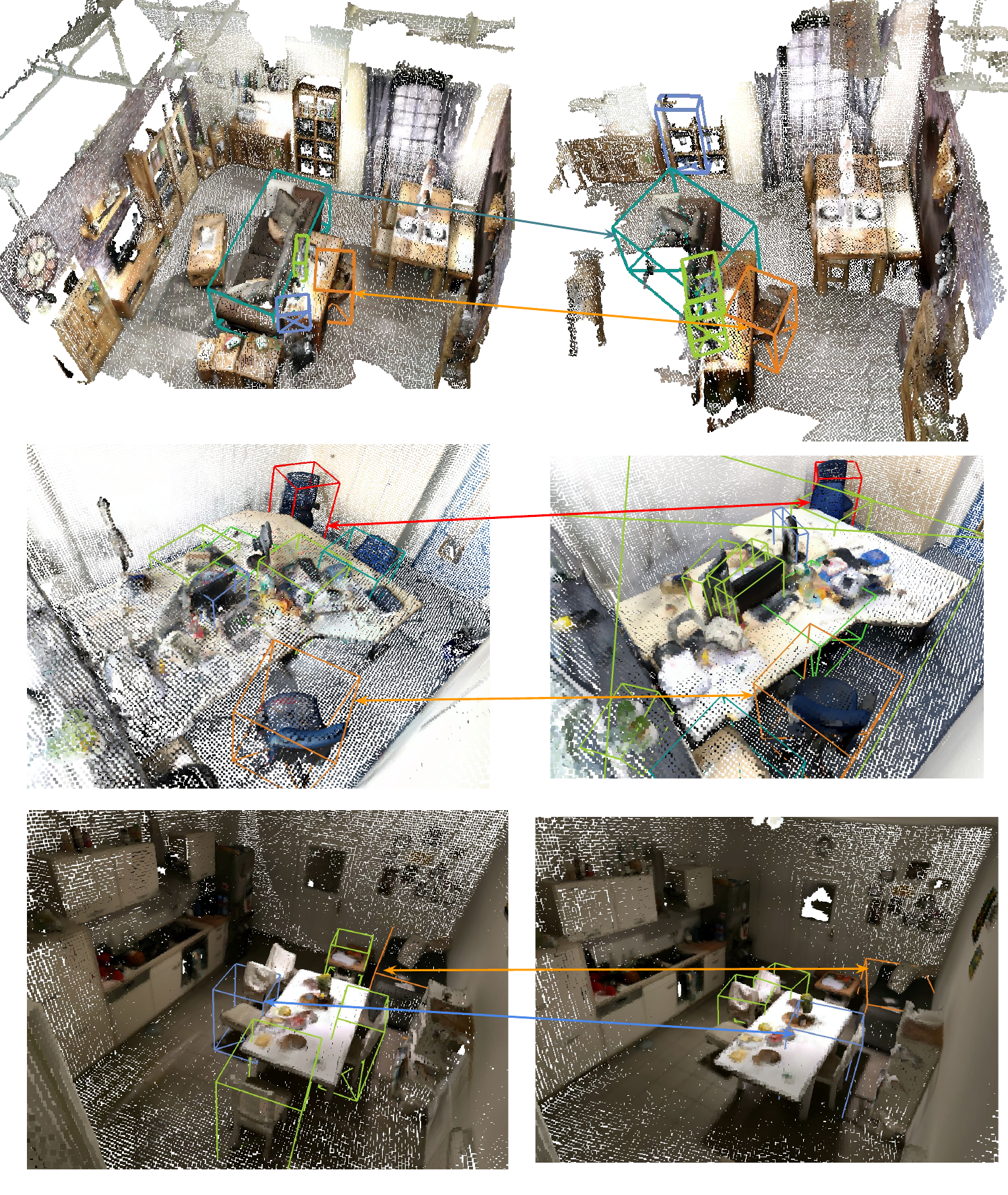}
    \caption{3D-SMNet qualitative results on RIO scenes \cite{wald2019rio}}
    \label{fig:3d_smnet_rio_results_supp}
\end{figure*}

\subsection{Baseline results comparison on RIO \cite{wald2019rio} when trained with different datasets.}
We report in Tab. \ref{tab:baselines_results_rio_dataaug} models performance comparison when trained with simulated data only (MP3D), on real data only (RIO) and on a combination of real and simulated data (RIO+MP3D). For the RIO+MP3D experiments we create a training set with the same amount of RIO and MP3D episodes. All evaluation numbers are reported on the validation set of RIO. Across each model we observe an increase in performances when the models are trained on both simulated and real data. This result strengthen the argument that using simulated episode as data augmentation helps boost the matching performances. We also notice that under each training regime 3D-SMNet outperforms all baselines. This result aligns with the outcome of the experiments on the Matterport3D and Replica datasets \cite{chang2017matterport3d,replica19arxiv}. Fig. \ref{fig:3d_smnet_rio_results_supp} shows qualitative results on the RIO datasets \cite{wald2019rio}.

\begin{table*}[t]
\centering
\caption{3D-SMNet and baseline matching results on the validation set of RIO \cite{wald2019rio} when trained with different datasets. 3D-SMNet performs best when trained jointly on real (RIO) and simulated (MP3D) episodes.}
\label{tab:baselines_results_rio_dataaug}
\renewcommand{\arraystretch}{1.15}
\resizebox{0.95\textwidth}{!}{
\begin{tabular}{l s c s c c c c }
\toprule
& & training dataset & & rank@1 & rank@5 & mAP & Acc \\ 
 H-L2 &  & N/A &  & 57.91 {\scriptsize $\pm$ 0.33 } & 92.90 {\scriptsize $\pm$ 0.00 } & 69.26 {\scriptsize $\pm$ 0.21 } & 45.07 {\scriptsize $\pm$ 0.33 } \\  
  H-M &  & N/A &  & 58.02 {\scriptsize $\pm$ 0.29 } & 91.33 {\scriptsize $\pm$ 0.06 } & 67.30 {\scriptsize $\pm$ 0.11 } & 44.47 {\scriptsize $\pm$ 0.37 } \\  
 \midrule
 H-1L &  & MP3D &  & 49.27 {\scriptsize $\pm$ 0.59 } & \textbf{100.00 {\scriptsize $\pm$ 0.00 }} & 70.85 {\scriptsize $\pm$ 0.35 } & 39.14 {\scriptsize $\pm$ 0.32 } \\  
 Sinkhorn &  & MP3D &  & 54.79 {\scriptsize $\pm$ 0.47 } & \textbf{100.00 {\scriptsize $\pm$ 0.00 }} & 70.00 {\scriptsize $\pm$ 0.31 } & 47.83 {\scriptsize $\pm$ 0.29 } \\  
 3D-SMNet &  & MP3D &  & 49.46 {\scriptsize $\pm$ 0.50 } & \textbf{100.00 {\scriptsize $\pm$ 0.00 }} & 70.00 {\scriptsize $\pm$ 0.29 } & 51.49 {\scriptsize $\pm$ 0.31 } \\  
 \arrayrulecolor{lightgray}
\midrule
 GTbox &  & MP3D &  & 60.74 {\scriptsize $\pm$ 0.20 } & 94.72 {\scriptsize $\pm$ 0.12 } & 75.42 {\scriptsize $\pm$ 0.14 } & 53.48 {\scriptsize $\pm$ 0.21 } \\
 \arrayrulecolor{black}
 \midrule
 H-1L &  & RIO &  & 54.20 {\scriptsize $\pm$ 0.56 } & 97.55 {\scriptsize $\pm$ 0.13 } & 71.60 {\scriptsize $\pm$ 0.35 } & 45.94 {\scriptsize $\pm$ 0.34 } \\  
 Sinkhorn &  & RIO &  & 54.80 {\scriptsize $\pm$ 0.51 } & 97.44 {\scriptsize $\pm$ 0.14 } & 71.99 {\scriptsize $\pm$ 0.31 } & 50.36 {\scriptsize $\pm$ 0.36 } \\  
 3D-SMNet &  & RIO &  & 51.68 {\scriptsize $\pm$ 0.49 } & \textbf{100.00 {\scriptsize $\pm$ 0.00 }} & 71.36 {\scriptsize $\pm$ 0.28 } & 51.40 {\scriptsize $\pm$ 0.29 } \\
 \arrayrulecolor{lightgray}
 \midrule
 GTbox &  & RIO &  & 61.32 {\scriptsize $\pm$ 0.19 } & 92.29 {\scriptsize $\pm$ 0.13 } & 74.04 {\scriptsize $\pm$ 0.14 } & 53.42 {\scriptsize $\pm$ 0.21 } \\ 
 \arrayrulecolor{black}
 \midrule
 H-1L &  & RIO + MP3D &  & 56.87 {\scriptsize $\pm$ 0.48 } & 97.33 {\scriptsize $\pm$ 0.13 } & 73.52 {\scriptsize $\pm$ 0.29 } & 47.36 {\scriptsize $\pm$ 0.30 } \\  
 Sinkhorn &  & RIO + MP3D &  & 54.82 {\scriptsize $\pm$ 0.44 } & \textbf{100.00 {\scriptsize $\pm$ 0.00 }} & 72.90 {\scriptsize $\pm$ 0.30 } & 54.75 {\scriptsize $\pm$ 0.32 } \\  
 3D-SMNet &  & RIO + MP3D &  & \textbf{62.76 {\scriptsize $\pm$ 0.51 }} & 97.38 {\scriptsize $\pm$ 0.14 } & \textbf{76.53 {\scriptsize $\pm$ 0.31 }} & \textbf{61.13 {\scriptsize $\pm$ 0.33 }} \\  
 \arrayrulecolor{lightgray}
 \midrule
 GTbox &  & RIO + MP3D &  & 67.05 {\scriptsize $\pm$ 0.22 } & 94.68 {\scriptsize $\pm$ 0.11 } & 77.92 {\scriptsize $\pm$ 0.16 } & 60.43 {\scriptsize $\pm$ 0.22 } \\ 
 \arrayrulecolor{black}
\bottomrule
\end{tabular}}

\end{table*}

\subsection{Limitations}
Our approach is fundamentally limited by the current state of the art in 3D detection as it serves as the foundation of our method. Whether training end-to-end models for this problem instead could be beneficial is not examined in this work. Our evaluation setting is also limited by the availability of 3D assets for insertion and the variety of 3D scenes to serve as environments. Choices we've made in selecting the objects and settings naturally result in a closed-world problem whereas object re-identification in real homes might face many new objects.
Such research may be applicable to building a new generation of home assistant. One potential negative impact that stands out from these applications is related to the privacy aspect. Potential users would need to be comfortable having a home robot or a pair of smart glasses that observe and create maps and representations of indoor homes

\end{document}